%% file: main.tex
\documentclass[10pt,twocolumn,letterpaper]{article}

\usepackage{cvpr}              % To produce the CAMERA-READY version

\definecolor{cvprblue}{rgb}{0.21,0.49,0.74}
\usepackage[pagebackref,breaklinks,colorlinks,allcolors=cvprblue]{hyperref}

\usepackage{colortbl}
\usepackage{textcomp}
\usepackage{gensymb}

\def\FL{\textrm{FL}}
\def\QFL{\textrm{QFL}}
\def\BCFL{\textrm{BCFL}}
\def\GC{\textrm{GC}}
\def\pt{p_\textrm{t}}
\def\at{\alpha_\textrm{t}}
\def\ac{\alpha_\textrm{c}}
\def\hP{\hat{P}}
\def\lmatch{{\cal L}_{\rm match}}
\def\MD{\textrm{MD}}
\def\CP{\textrm{CP}}
\def\CAS{\textrm{CAS}}
\def\person{\textrm{person}}
\def\skis{\textrm{skis}}
\def\bottle{\textrm{bottle}}

\newcommand{\eqnnm}[2]{\begin{equation}\label{eq:#1}#2\end{equation}\ignorespaces}
\newcommand{\ntb}{\noindent \textbf}
\newcommand{\vtb}{\vspace{3pt}\noindent \textbf}
\newcommand{\greyrule}{\arrayrulecolor{black!30}\midrule\arrayrulecolor{black}}
\newcommand\blfootnote[1]{% 
    \begingroup 
    \renewcommand\thefootnote{}\footnote{#1}% 
    \addtocounter{footnote}{-1}% 
    \endgroup 
}
\title{OCDet: Object Center Detection via Bounding Box-Aware Heatmap Prediction on Edge Devices with NPUs}

\author{
Chen Xin$^{1,2,\dagger}$
\quad
Thomas Motz$^{2}$
\quad
Andreas Hartel$^{2}$
\quad
Enkelejda Kasneci$^{1}$
\\[1pt]
$^{1}$ Technical University of Munich
\quad
$^{2}$ Liebherr-Electronics and Drives GmbH
\vspace{-4pt}
}

\begin{document}
\maketitle
\blfootnote{$\dagger$: corresponding author. Email: \href{chen.xin@tum.de}{chen.xin@tum.de}}
\input{sec/0_abstract}    
\input{sec/1_intro}
\input{sec/2_related}
\input{sec/3_method}
\input{sec/4_experiments}

\input{sec/5_conclusion}
{
    \small
    \bibliographystyle{ieeenat_fullname}
    \bibliography{main}
}
\input{sec/6_suppl}

\end{document}

%% file: sec/0_abstract.tex
\begin{abstract}

Real-time object localization on edge devices is fundamental for numerous applications, ranging from surveillance to industrial automation. Traditional frameworks, such as object detection, segmentation, and keypoint detection, struggle in resource-constrained environments, often resulting in substantial target omissions. To address these challenges, we introduce \textbf{OCDet}, a lightweight \textbf{O}bject \textbf{C}enter \textbf{Det}ection framework optimized for edge devices with NPUs. OCDet predicts heatmaps representing object center probabilities and extracts center points through peak identification. Unlike prior methods using fixed Gaussian distribution, we introduce Generalized Centerness (GC) to generate ground truth heatmaps from bounding box annotations, providing finer spatial details without additional manual labeling. Built on NPU-friendly Semantic FPN with MobileNetV4 backbones, OCDet models are trained by our Balanced Continuous Focal Loss (BCFL), which alleviates data imbalance and focuses training on hard negative examples for probability regression tasks. Leveraging the novel Center Alignment Score (CAS) with Hungarian matching, we demonstrate that OCDet consistently outperforms YOLO11 in object center detection, achieving up to 23\% higher CAS while requiring 42\% fewer parameters, 34\% less computation, and 64\% lower NPU latency. When compared to keypoint detection frameworks, OCDet achieves substantial CAS improvements up to 186\% using identical models. By integrating GC, BCFL, and CAS, OCDet establishes a new paradigm for efficient and robust object center detection on edge devices with NPUs. The code is released at \href{https://github.com/chen-xin-94/ocdet}{https://github.com/chen-xin-94/ocdet}.

\end{abstract} 

%% file: sec/1_intro.tex
\section{Introduction}
\label{sec:intro}

\begin{figure}[t]
    \vspace{-0.5cm}
    \centering
    \includegraphics[width=1.0\linewidth]{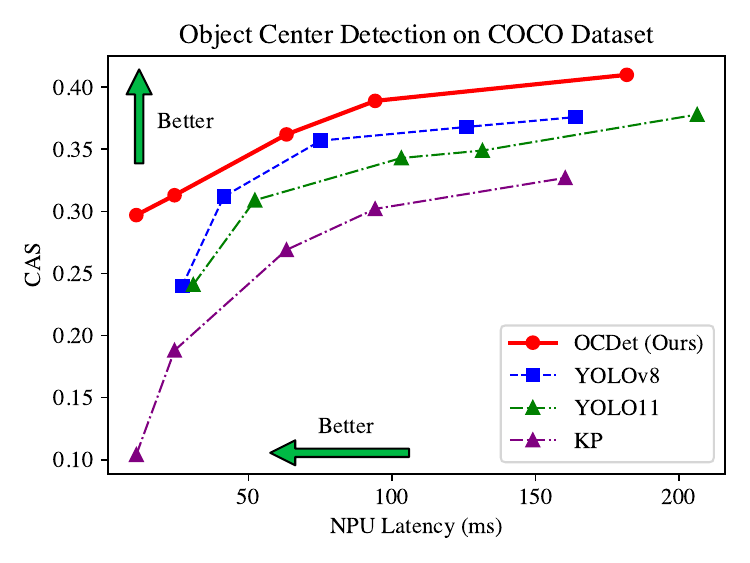}
    \vspace{-0.8cm}
    \caption{Performance comparison between our proposed OCDet framework and state-of-the-art real-time object detectors YOLOv8 and YOLO11, as well as a standard keypoint detection framework (KP) using identical model configurations. Results demonstrate OCDet’s Pareto dominance in latency-accuracy trade-offs.}
    \vspace{-0.3cm}
    \label{fig:cas_latency}
\end{figure}

Automated localization of target objects is a critical task in various computer vision applications such as surveillance \cite{sreenuIntelligentVideoSurveillance2019, duongDeepLearningBasedAnomaly2023}, autonomous driving \cite{michaelisBenchmarkingRobustnessObject2020, fengReviewComparativeStudy2022}, and industrial automation \cite{sagaCaseStudyPerformance2020, zhouLearningbasedObjectDetection2022}. Over the years, extensive research has focused on three primary frameworks to address this challenge: object detection for bounding box regression, segmentation for pixel-level delineation, and keypoint detection for target point identification. Each framework has made significant progress in object identification and localization with 2D images.

Despite these advancements, prior approaches struggle in ubiquitous real-world applications that require only coarse localization but demand real-time performance with minimal computational resources. In such scenarios, the primary objective is to detect the presence or approximate location of objects, typically their centers, rather than detailed boundary delineation. Each of the above-mentioned conventional frameworks has specific drawbacks in this context. Keypoint detection frameworks \cite{xiaoSimpleBaselinesHuman2018, lawCornerNetDetectingObjects2019, duanCenterNetKeypointTriplets2019, zhouObjectsPoints2019, caoOpenPoseRealtimeMultiPerson2019, wangDeepHighResolutionRepresentation2020, chengHigherHRNetScaleAwareRepresentation2020, liSimCCSimpleCoordinate2022, chenCascadedPyramidNetwork2018, newellAssociativeEmbeddingEndEnd2017, gengBottomHumanPoseEstimation2021, kreissPifPafCompositeFields2019}, primarily developed for human pose estimation, depend on intricate manual labeling yet simplified ground truth generation methods, limiting its applicability to general center detection tasks. Object detection approaches \cite{renFasterRCNNRealTime2016, carionEndEndObjectDetection2020, jocherUltralyticsYOLOv82023, lvDETRsBeatYOLOs2023, zongDETRsCollaborativeHybrid2023, jocherUltralyticsYOLO112024}, on the other hand, prioritize precise bounding box prediction over avoiding object omissions, which is counterproductive for applications that require high recall, such as surveillance. Lastly, segmentation methods \cite{longFullyConvolutionalNetworks2015, zhaoPyramidSceneParsing2017, chenDeepLabSemanticImage2017, heMaskRCNN2018, xieSegFormerSimpleEfficient2021, chengMaskedattentionMaskTransformer2022}, while offering pixel-level precision, are computationally intensive and thus impractical for deployment on edge devices.

To address these limitations, we introduce \textbf{OCDet}, a lightweight \textbf{O}bject \textbf{C}enter \textbf{Det}ection framework, tailored for edge devices with NPUs. We transform the coarse localization demand to the center detection task and define the object center as the central point of its bounding box. Instead of direct regression as in most object detection methods, OCDet predicts heatmaps encoding object center probabilities and then extracts center coordinates via peak identification. Unlike keypoint detection approaches, which adopt fixed Gaussian heatmaps centered at target points, our ground truth heatmaps are produced by the proposed Generalized Centerness (GC) based on bounding box annotations. With bounding box-aware GC, OCDet works seamlessly with object detection datasets \cite{everinghamPascalVisualObject2015, linMicrosoftCOCOCommon2015, shaoObjects365LargeScaleHighQuality2019} and integrates smoothly with automated bounding box labeling approaches \cite{xinDARTAutomatedEndEnd2024,fengInstaGenEnhancingObject2024} for novel categories, eliminating the need for extra manual annotation. Optimized for NPUs, OCDet models are built on Semantic FPN \cite{kirillovPanopticFeaturePyramid2019} and MobileNetV4 backbones \cite{qinMobileNetV4UniversalModels2024}. To enhance model training on zero-dominant GC heatmaps, we introduce Balanced Continuous Focal Loss (BCFL) to handle data imbalance while emphasizing learning on hard negatives for probability regression tasks with skewed target distributions. Lastly, we present a novel Center Alignment Score (CAS) to validate OCDet's effectiveness. CAS applies the Hungarian matching to the predicted and ground truth centers and internalizes the concept of precision and recall into Matched Discrepancy (MD) and Cardinality Penalty (CP). The resulting CAS, as a normalized distance metric, provides a holistic measure of model performance and is broadly applicable to center detection tasks on general object detection datasets.

Extensive experiments on the COCO dataset \cite{linMicrosoftCOCOCommon2015} demonstrate OCDet’s significant improvements over traditional frameworks. As illustrated in \cref{fig:cas_latency}, OCDet achieves substantial improvements of 23\% / 1\% / 6\% / 11\% / 8\% in CAS compared to the latest real-time object detector YOLO11 \cite{jocherUltralyticsYOLO112024} across all five model variants while requiring 42\% / 83\% / 61\% / 25\% / 61\% fewer parameters, 34\% / 65\% / 69\% / 30\% / 71\% less computation, and achieving 64\% / 54\% / 39\% / 28\% / 22\% lower NPU latency. In comparison with a standard keypoint detection framework (KP), OCDet demonstrates superior performance under identical model configurations, delivering remarkable CAS grains by 186\% / 66\% / 35\% / 29\% / 25\%, respectively. These achievements highlight OCDet as a powerful and efficient solution for object center detection on edge devices with NPUs.

The contributions of this paper are summarized as follows: 
(1) We introduce OCDet, a novel \textbf{O}bject \textbf{C}enter \textbf{Det}ection framework, that predicts heatmaps generated by the proposed bounding box-aware Generalized Centerness (GC), offering an efficient solution to real-world applications that require only coarse localization but demand real-time performance with minimal computational resources. 
(2) OCDet is optimized for NPU, utilizing models based on Semantic FPN with MobileNetV4 backbones to achieve accurate and robust object center detection. 
(3) Model training incorporates our proposed Balanced Continuous Focal Loss (BCFL) to handle data imbalance and focus on hard negative examples for probability regression tasks with skewed target distributions, resulting in significant performance improvements. 
(4) We define a new metric, Center Alignment Score (CAS), which leverages Hungarian matching and integrates precision and recall into a unified, normalized distance-based measure for evaluating object center detection performance.
(5) Through extensive experiments, we show that OCDet consistently outperforms the latest real-time object detector YOLO11, achieving up to 23\% higher CAS while reducing parameters by 42\%, computation by 34\%, and NPU latency by 64\%. OCDet also demonstrates a CAS increase of up to 186\% over a standard keypoint detection framework (KP) in identical model configurations.

\begin{figure*}[ht]
    \vspace{-0.7cm}
    \centering
    \includegraphics[width=0.85\linewidth]{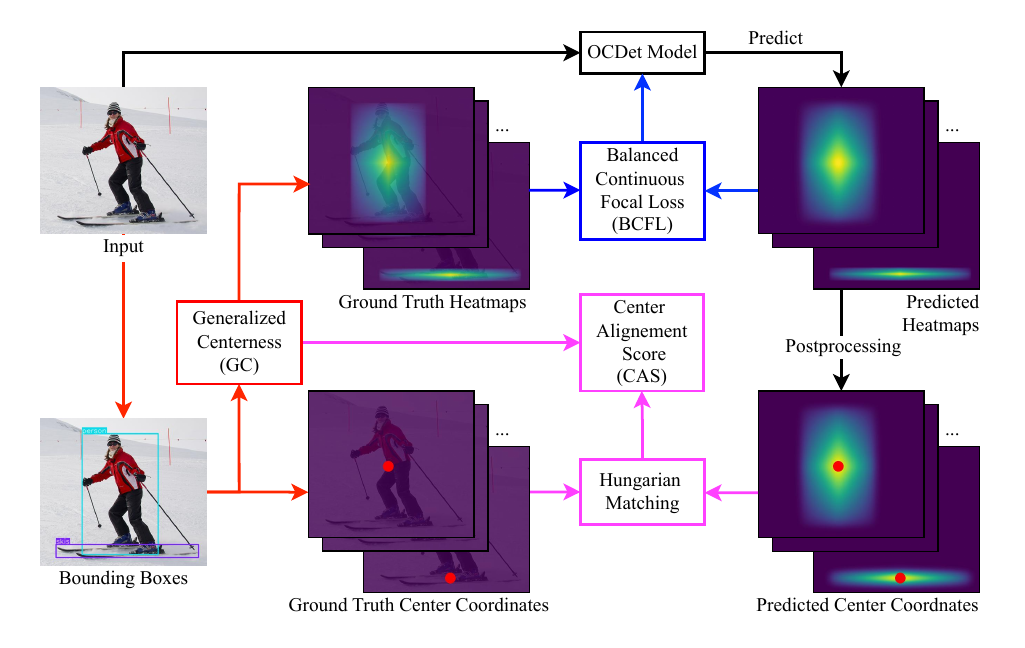}
    \vspace{-0.5cm}
    \caption{Illustration of our proposed OCDet, an Object Center Detection framework, comprising: \textcolor{red}{ground truth generation} leveraging Generalized Centerness (GC), \textcolor{blue}{model training} through Balanced Continuous Focal Loss (BCFL), \textcolor{black}{inference} with NPU-oriented OCDet models, and \textcolor{magenta}{evaluation} using Hungarian matching and Center Alignment Score (CAS).}
    \vspace{-0.3cm}   
    \label{fig:ocd}
\end{figure*}

%% file: sec/2_related.tex
\section{Related Work}
\label{sec:related}
 
\ntb{Keypoint Detection} involves simultaneously identifying and localizing target points of an object, primarily designed for human pose estimation. Heatmap-based approaches currently dominate the field. These methods utilize powerful backbones \cite{newellStackedHourglassNetworks2016, caoOpenPoseRealtimeMultiPerson2019, wangDeepHighResolutionRepresentation2020,yuLiteHRNetLightweightHighResolution2021,chengHigherHRNetScaleAwareRepresentation2020} to predict probability heatmaps and retrieve keypoint coordinates via postprocessing \cite{newellAssociativeEmbeddingEndEnd2017, xiaoSimpleBaselinesHuman2018,chenCascadedPyramidNetwork2018,caoOpenPoseRealtimeMultiPerson2019,liSimCCSimpleCoordinate2022}. However, previous works \cite{xiaoSimpleBaselinesHuman2018, lawCornerNetDetectingObjects2019,duanCenterNetKeypointTriplets2019, zhouObjectsPoints2019, caoOpenPoseRealtimeMultiPerson2019, wangDeepHighResolutionRepresentation2020, chengHigherHRNetScaleAwareRepresentation2020,liSimCCSimpleCoordinate2022,chenCascadedPyramidNetwork2018,newellAssociativeEmbeddingEndEnd2017,gengBottomHumanPoseEstimation2021, kreissPifPafCompositeFields2019} adopt a rather simplistic approach to generating the ground truth maps: employing a 2D Gaussian distribution with a constant standard deviation centered at each keypoint. The fixed Gaussian distribution fails to capture the diverse shapes and sizes of various objects, leading to inferior performance. In contrast, we propose Generalized Centerness (GC) to generate bounding box-aware heatmaps, providing finer granularity and increased control over the ground truth heatmap without the need for additional manual labeling.

\vtb{Object detection} aims to accurately classify and locate target objects with bounding boxes. The YOLO family \cite{redmonYouOnlyLook2016, redmonYOLO9000BetterFaster2016, redmonYOLOv3IncrementalImprovement2018, bochkovskiyYOLOv4OptimalSpeed2020, jocherYOLOv5Ultralytics2020, liYOLOv6SingleStageObject2022, wangYOLOv7TrainableBagfreebies2022, jocherUltralyticsYOLOv82023, wangYOLOv9LearningWhat2024, wangYOLOv10RealTimeEndEnd2024, jocherUltralyticsYOLO112024} establishes a real-time framework where each cell on the feature map regresses its nearby center coordinates and bounding box edges. Building on this framework, subsequent works refine the building block \cite{redmonYOLOv3IncrementalImprovement2018, bochkovskiyYOLOv4OptimalSpeed2020, jocherYOLOv5Ultralytics2020, liYOLOv6SingleStageObject2022, wangYOLOv7TrainableBagfreebies2022, jocherUltralyticsYOLOv82023, wangYOLOv9LearningWhat2024, wangYOLOv10RealTimeEndEnd2024, jocherUltralyticsYOLO112024}, ground truth representation \cite{redmonYOLOv3IncrementalImprovement2018, bochkovskiyYOLOv4OptimalSpeed2020,jocherUltralyticsYOLOv82023}, loss function \cite{bochkovskiyYOLOv4OptimalSpeed2020, wangYOLOv7TrainableBagfreebies2022, wangYOLOv9LearningWhat2024, wangYOLOv10RealTimeEndEnd2024}, postprocessing \cite{bochkovskiyYOLOv4OptimalSpeed2020, wangYOLOv10RealTimeEndEnd2024}, and so on. From the perspective of center detection, object detection can be seen as a specialized form of regression-based keypoint localization \cite{toshevDeepPoseHumanPose2014,carreiraHumanPoseEstimation2016, papandreouPersonLabPersonPose2018, dijkstraCentroidNetDeepNeural2019}. In this context, the target is exclusively the center point, while the predicted bounding box edges serve to remove duplicates and select promising candidates. The primary distinction between YOLO and OCDet lies in the representation of ground truth. While YOLO represents each object bounding box with four scalars (center coordinates, height, and weight), OCDet uses the proposed GC to produce a dense heatmap that spans the entire bounding box region, providing finer spatial details. Our experiments demonstrate that OCDet surpasses state-of-the-art YOLOv8 and YOLOv11 in object center detection with significant improvements in recall and our proposed CAS metric.

\vtb{Segmentation} generates a 0-1 mask to classify each pixel as an object or the background. Similar to OCDet, modern image segmentation utilizes neural networks for pixel-wise prediction. However, while segmentation demands accurate prediction for each pixel, OCDet primarily aims to achieve high prediction accuracy within the central region of an object, a goal achieved by our proposed BCFL. Consequently, OCDet eliminates the need for elaborate structures typical of segmentation models \cite{longFullyConvolutionalNetworks2015, zhaoPyramidSceneParsing2017, chenDeepLabSemanticImage2017,heMaskRCNN2018, xieSegFormerSimpleEfficient2021,chengMaskedattentionMaskTransformer2022}. In contrast, it adopts lightweight MobileNetV4 backbones and the straightforward Semantic FPN architecture, enabling real-time performance on resource-constrained NPUs.

%% file: sec/3_method.tex
\section{Method}
\label{sec:method}

We propose \textbf{OCDet}, a lightweight \textbf{O}bject \textbf{C}enter \textbf{Det}ection framework. As illustrated in \cref{fig:ocd}, OCDet first generates ground truth heatmaps based on Generalized Centerness (GC), encoding the likelihood of each pixel being an object center (\cref{sec:gc}). To predict these heatmaps, we design OCDet models with NPU-friendly Semantic FPN and MobileNetV4 (\cref{sec:model}). These models are trained with Balanced Continuous Focal Loss (BCFL), which addresses data imbalance and directs learning attention to hard negatives (\cref{sec:bcfl}). To verify OCDet's performance, we propose the Center Alignment Score (CAS) in \cref{sec:cas}, a normalized distance-based metric utilizing Hungarian matching (\cref{sec:hungarian}).

%-------------------------------------------------------------------------
\subsection{Generalized Centerness}
\label{sec:gc}
The concept of Centerness is first introduced in FCOS \cite{tianFCOSFullyConvolutional2019}. It depicts the normalized distance from a point inside the bounding box to the box center. Our Generalized Centerness (GC) introduces an adaptive mechanism, enabling it to dynamically adjust to diverse object shapes. Given a location $(x,y)$, let $l(x)$, $r(x)$, $t(y)$, $b(y)$ represent the distance from this point to the left, right, top, and bottom edges of its bounding box, GC is defined as:
\eqnnm{gc}{\GC(x,y) = \left( \frac{ \min(l, r)}{ \max(l, r)} \right) ^{\eta} \times \left( \frac{ \min(t, b)}{ \max(t, b)} \right) ^{\phi},}
where $\eta$ and $\phi$ are hyperparameters that control the shape of the GC heatmap. As $\eta$ and $\phi$ increase, the GC heatmap becomes more concentrated around the center. Decreasing $\eta$ or $\phi$ causes the concentration to spread along the horizontal or vertical axis, respectively. When $\eta=\phi=0.5$, GC reduces to Centerness. \cref{fig:gc} in the supplementary material illustrates GC maps generated under various settings of $\eta$ and $\phi$, while \cref{sec:ablation} provides an in-depth analysis of the effects of different $\eta$ and $\phi$ combinations.
 
Unlike FCOS, where Centerness is regarded as a normalized distance and used solely as a weighting factor, our OCDet leverages GC to generate ground truth probability heatmaps. Each point within a bounding box is assigned a ground truth value based on \cref{eq:gc}, while points outside the box receive a value of zero. Since GC ranges from [0,1], each pixel's assigned value can be directly interpreted as its probability of being an object center, which is defined as the midpoint of a bounding box.

Ground truth GC heatmaps are generated individually for each class. Consequently, for the COCO dataset, $\GC(x,y)$ is applied separately to produce 80 class-wise heatmaps for OCDet models to predict. To resolve ambiguity in overlapping bounding box regions within the same category, we assign the highest $\GC$ value to a location when it falls within multiple bounding boxes. This strategy prioritizes the integrity of the heatmap for smaller objects, as the overlapping region is more likely to be closer to the center of a small instance, resulting in a higher GC value. 
%-------------------------------------------------------------------------
\subsection{Model}
\label{sec:model}

\ntb{NPU-oriented.} With the ground truth GC heatmaps established, we developed the OCDet model to predict them. We focus on lightweight architectures and backbones that are well-suited to both the prediction task and the hardware constraints of NPUs. NPUs come with particular challenges. While traditional modules such as standard, depthwise, and pointwise convolutions, along with ReLU and BatchNorm, demonstrate high efficiency and compatibility, other widely used modules such as GELU \cite{hendrycksGaussianErrorLinear2023}, H-Swish \cite{howardSearchingMobileNetV32019}, SE-blocks \cite{huSqueezeExcitationNetworks2019} in modern CNNs, as well as LayerNorm \cite{baLayerNormalization2016} and MHSA \cite{vaswaniAttentionAllYou2017} in transformer architectures are not well optimized for NPUs. Our tests confirm that these modules introduce latency and degrade overall performance, consistent with the findings presented in \cite{vasuMobileOneImprovedOne2023,qinMobileNetV4UniversalModels2024}. Therefore, we focus exclusively on simple architectures and standard convolutional modules. 

\vtb{Architecture and backbone.} Considering NPU's constraints, we adopt the Semantic FPN \cite{kirillovPanopticFeaturePyramid2019} architecture with MobileNetV4 \cite{qinMobileNetV4UniversalModels2024} backbones for our OCDet models. These models (except for the smallest variant OCDet-N, see \cref{app:imp}) extract features from the last four levels of the backbone's feature pyramid to produce an output at 1/4 of the image resolution. The straightforward design of FPN and the standardized modules in MobileNetV4 deliver impressive speed performance on NPU. For a more detailed model description, please refer to \cref{app:models}.

\vtb{Peak Identification.} Given predicted heatmaps by the OCDet models, we identify the peaks to obtain the output center coordinates. We apply non-maximum suppression (NMS) following prior works \cite{lawCornerNetDetectingObjects2019, papandreouPersonLabPersonPose2018, chenCascadedPyramidNetwork2018} and filter with minimum probability and distance constraints to reduce false positives and redundant centers (see \cref{app:ablation}).

%-------------------------------------------------------------------------
\subsection{Balanced Continuous Focal Loss}
\label{sec:bcfl}

We propose Balanced Continuous Focal Loss (BCFL) to train OCDet models. BCFL addresses the data imbalance issue and simultaneously focuses training on hard negative examples, making it well-suited for continuous probability regression tasks characterized by skewed target distributions, like object center detection.
 
We introduce BCFL by first examining the Focal Loss (FL) \cite{linFocalLossDense2018} for binary classification:
\begin{align}
    \begin{split}
        \FL(\pt) &= - \at(1 - \pt)^\gamma \log (\pt), \\
        \text{where} \quad
        \pt &=
        \begin{cases} 
            p, & \text{if } y = 1, \\ 
            1 - p, & \text{otherwise},
        \end{cases} \\
        \text{and} \quad
        \alpha_t &=
        \begin{cases} 
            \alpha, & \text{if } y = 1, \\ 
            1 - \alpha, & \text{otherwise}.
        \end{cases}
    \end{split}
    \label{eq:fl}
\end{align}
Here, $y \in \{\pm 1\}$ represents the ground-truth class, $p \in [0,1]$ denotes the model’s predicted probability for the positive class $y=1$, and $\alpha \in [0,1]$ is an optional weighting factor to address the class imbalance. Compared to standard cross entropy loss $-\log (\pt)$, FL introduces a modulating factor $(1-\pt)^\gamma$ to direct learning attention to hard negatives.

The Quality Focal Loss (QFL) \cite{liGeneralizedFocalLoss2020a} extends FL from classification with discrete labels $y \in \{\pm 1\}$ to regression tasks with continuous target $y \in [0,1]$ by (1) expanding the cross entropy part $-\log (\pt)$ to its complete version using continuous target $y$ and (2) generalizing the scaling factor $(1-\pt)$ to the absolute distance between the prediction $p$ and its continuous label $y$:
\eqnnm{qfl}{\QFL(p) = - |y - p|^{\gamma} \left[ (1 - y) \log(1 - p) + y \log(p) \right].}

Although QFL inherits the hard negative mining from FL, it does not address data imbalance, rendering it unsuitable for heatmap-based object center detection due to the prevalence of zero-valued points in the ground truth. Furthermore, the importance of different regions in center detection tasks is not uniform. Accurate probability estimates in central regions are particularly crucial, as they directly impact downstream peak identification and ultimately influence the accuracy of the final predicted center coordinates.

\begin{figure}[h]
    \centering
    \vspace{-0.1cm}
    \begin{subfigure}{0.5\columnwidth}
        \centering
        \includegraphics[width=\linewidth]{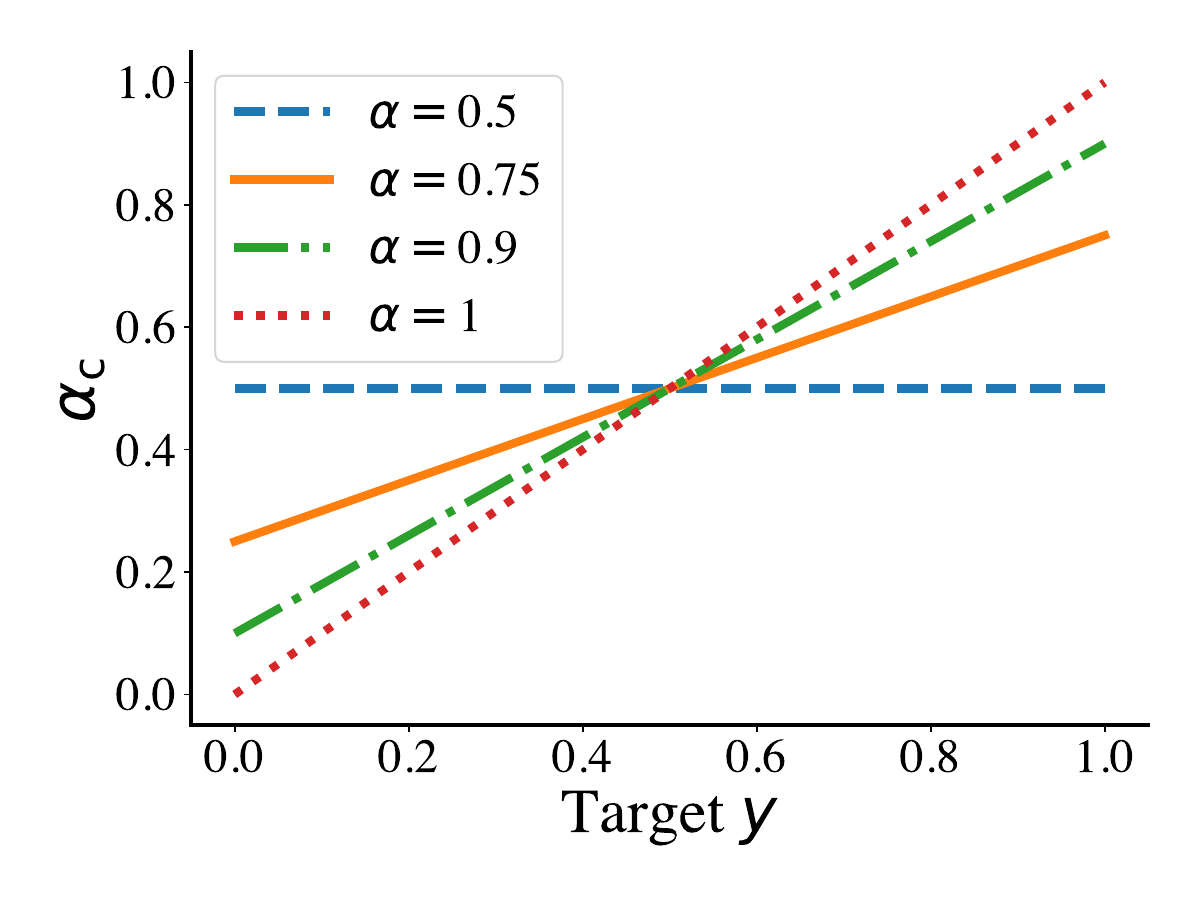}
        \vspace{-0.5cm}
        \caption{$\ac(y)$ under various $\alpha$ values}
    \end{subfigure}%
    \hfill
    \begin{subfigure}{0.5\columnwidth}
        \centering
        \includegraphics[width=\linewidth]{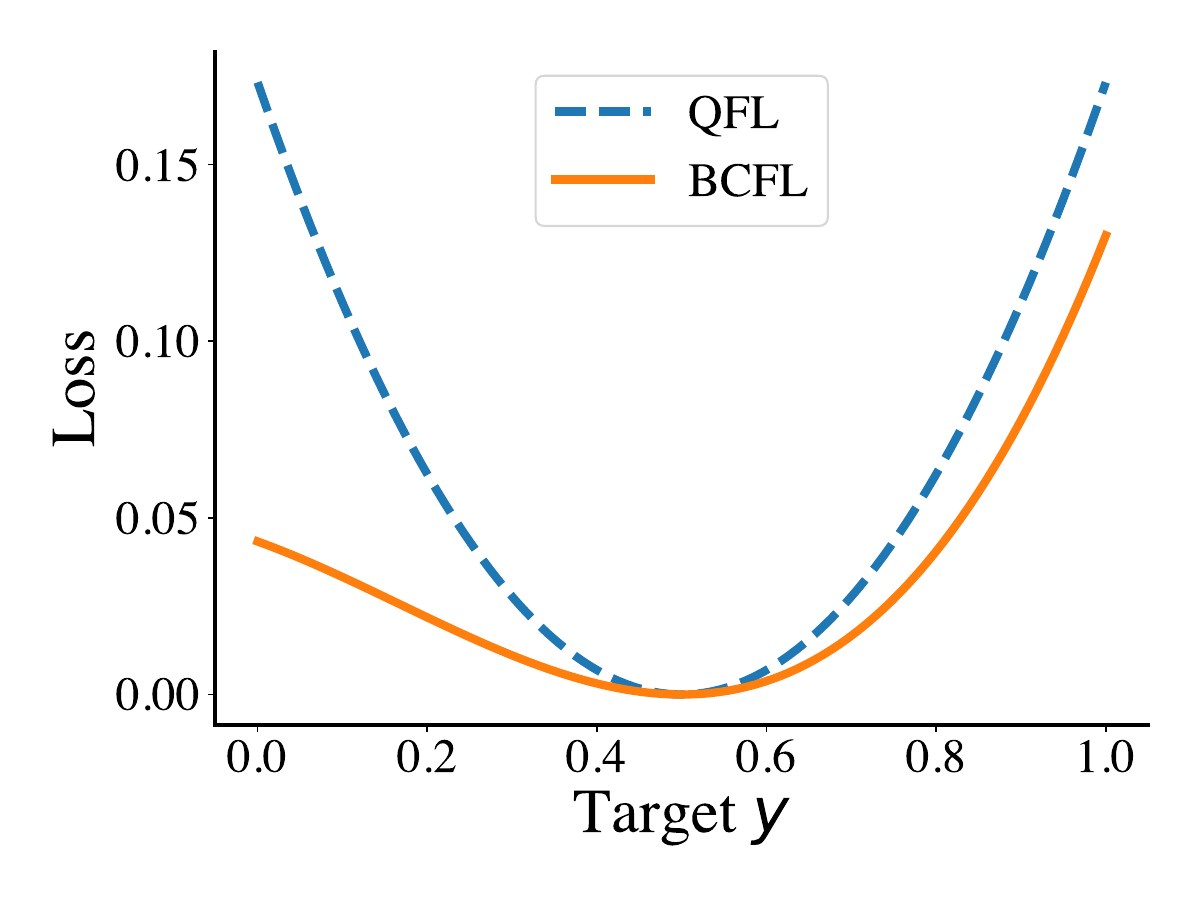}
        \vspace{-0.5cm}
        \caption{BCFL vs QFL}
    \end{subfigure}
    \vspace{-0.5cm}
    \caption{Illustration of (a) $\ac(y)$ given a certain $\alpha$ and (b) BCFL vs QFL under $\alpha=0.75$ across various target values $y \in [0,1]$.}
    \vspace{-0.2cm}
    \label{fig:bcfl}
\end{figure}

To overcome these challenges, we propose Balanced Continuous Focal Loss (BCFL), which dynamically adjusts sample weights based on target values during loss computation. This strategy effectively addresses data imbalance and directs training attention to central regions with higher target values simultaneously. Inspired by the $\at$-balanced focal loss in \cref{eq:fl}, BCFL adapts QFL with a novel adaptive weighting factor $\ac(y)$, calibrated dynamically based on the target probability value: 
\begin{align}
    \BCFL(p) = &\ac(y) |y - p|^{\gamma} \left[ (1 - y) \log(1 - p) + y \log(p) \right], \notag \\
    \text{where} \quad &\ac(y) = \alpha y + (1-\alpha)(1-y). \label{eq:bcfl}
\end{align}
Here, $\ac(y)$ is a convex combination of two probabilities that the given location is a center ($y$) and that it is not a center ($1 - y$). As shown in \cref{fig:bcfl} (a), setting $\alpha$ within the range $0.5 \leq \alpha \leq 1$ serves to reduce weight for distant points with lower target values and emphasize center regions with higher target values. In practice, $\alpha$ is set to the inverse class frequency. As $\alpha$ increases, higher target values progressively gain larger $\ac(y)$, while lower target values are increasingly suppressed. This trend is further highlighted in \cref{fig:bcfl} (b), where BCFL considerably reduces the loss for lower target values while preserving relatively larger losses for high-value targets, thereby accentuating points near the object centers.

 %-------------------------------------------------------------------------
\subsection{Hungarian Matching}
\label{sec:hungarian}

We obtain the final predicted center points by applying peak identification (\cref{sec:model}) to the heatmaps generated by OCDet models trained with BCFL (\cref{sec:bcfl}). In preparation for model evaluation, we utilize the Hungarian algorithm \cite{kuhnHungarianMethodAssignment1955} to find an optimal bipartite matching between predicted $\{\hP_i\}_{i=1}^{N}$ and ground truth centers $\{P_i\}_{i=1}^{G}$. To determine matching points, we search for a permutation $\sigma \in \Sigma_{N}$ (assuming $N>G$) with the lowest matching cost:
\eqnnm{hungarian}{\hat{\sigma} = \arg\min_{\sigma \in \Sigma_N} \sum_{i=1}^{} \lmatch({P_i, \hP_{\sigma(i)}}).}
Here, the matching cost $\lmatch$ considers both the distance and probability ($\GC$) difference between a pair of ground truth ($P_i$) and predicted point ($\hP_j$). It's defined as:
\eqnnm{match}{\lmatch(P_i, \hP_j) = \lambda \| P_i - \hP_j \|_2 + \mu |\GC(P_i) - p(\hP_j)|,}
where the contribution of distance and probability difference cost is controlled by $\lambda$ and $\mu$, which are both set to 1 in this work. Given the cost matrix for pairwise combinations between ground truth and predictions, the Hungarian algorithm can compute the optimal assignment efficiently.

%-------------------------------------------------------------------------
\subsection{Center Alignment Score}
\label{sec:cas}

To validate the effectiveness of the OCDet model, we propose a novel metric, Center Alignment Score (CAS).

We first refine the matched points obtained through Hungarian matching (\cref{sec:hungarian}). For each ground truth center $P_m$, we assign a threshold as the distance to its box corner, defined as $D_i=0.5\sqrt{w_i^2+h_i^2}$. This value represents the maximum possible distance from the center to any point within the bounding box, essentially half of the box diagonal. If the actual distance between a matched pair exceeds this threshold, we reclassify the pair as unmatched.

For all remaining matched points $\mathcal{M}$ on image $i$, we normalize the coordinate distances of all matched pairs by the distance threshold $D_i$. The average of these normalized distances is termed as the Matched Discrepancy (MD):
\eqnnm{md}{\MD_i = \sum_{m=1}^{|\mathcal{M}|} \frac{\| P_m - \hP_m \|}{D_m}.}

To incorporate unmatched points into CAS, we assign each point a normalized distance of one. This can be understood as assigning each unmatched point a virtual bounding box. The loss for each unmatched point is then set as the distance from the center to the farthest corner of this box. When the distance is normalized by the distance threshold as in \cref{eq:md} for unit consistency, it yields a value of one. With this unit penalty for each unmatched point, we define the Cardinality Penalty (CP) as the larger count between unmatched predictions and unmatched ground truth points:
\eqnnm{cp}{\CP_i = \max \left( \operatorname{card}(P_i \setminus \mathcal{M}), \operatorname{card}(\hat{P}_i \setminus \mathcal{M}) \right),}
where $\operatorname{card}$ is cardinality and $\setminus$ denotes set difference.

With CP and MD, the final CAS score starts at the perfect score 1, from which the two penalty terms are averaged across all images and then subtracted:
\eqnnm{cas}{\CAS = 1-\CP-\MD = 1 -\sum\frac{ \CP_i }{ N_i }-\sum\frac{\MD_i}{ N_i }.}

CAS is broadly applicable to center detection tasks on any object detection dataset. Matched Distance (MD) and Coverage Precision (CP) naturally extend the concepts of precision and recall in binary classification to continuous measurements of localization accuracy. Low MD and CP indicate high precision and recall, respectively. Unlike COCO’s AP metric \cite{linMicrosoftCOCOCommon2015}, which relies on fixed IoU thresholds (typically [0.5:.05:0.95]) to discretize regression into classification for standard metric use, our proposed CAS internalizes the principles of precision and recall into a unified, normalized distance measure. CAS provides a continuous, scale-aware evaluation that avoids the abrupt performance changes induced by fixed thresholds, resulting in a more stable and consistent assessment of center detection performance across varying object scales.

%% file: sec/4_experiments.tex
\section{Experiments}
\label{sec:experiments}

In this section, we present experiments conducted on the proposed OCDet framework for object center detection tasks. The experimental setup is outlined in \cref{sec:setup} with further model and training details in \cref{app:imp}. Ablation studies on the proposed Generalized Centerness (GC) and Balanced Continuous Focal Loss (BCFL) are discussed in \cref{sec:ablation}, while additional studies on model design are provided in \cref{app:ablation}. Finally, we present a qualitative analysis and visualizations of the model outputs in \cref{app:vis}.

%-------------------------------------------------------------------------
\subsection{Experimental Setup}
\label{sec:setup}

\ntb{Dataset.} We conduct our experiments on the COCO dataset. Ground truth center points are derived from the bounding box annotations. The train2017 set serves for training and hyperparameter tuning, while the val2017 set is used for validation.

\vtb{Hardware.} All latency measurements are conducted on the NPU of i.MX 8M Plus, which delivers 2.3 TOPS. For NPU inference, models are converted to TensorFlow Lite format and quantized to uint8. 

\vtb{Model Details.} Due to the limited resources on edge devices, the model input is resized to $320 \times 320$. Based on MobilenetV4 and Semantic FPN, we build a series of OCDet models with 5 variants: N/S/M/L/X, spanning model sizes from Nano to eXtra-large. Further details on each model configuration are provided in \cref{app:imp}.

\vtb{Training Details.} OCDet models are trained for 24 epochs with a batch size of 64 using the AdamW optimizer \cite{loshchilovDecoupledWeightDecay2017}. The initial learning rate ranges from $7 \times 10^{-4}$ to $6 \times 10^{-3}$ depending on model size and follows a cosine decay to zero after a warm-up phase.  We use BCFL as the loss function with $\gamma = 2$ and $\alpha = 0.984$, derived from the class inverse frequency with probability threshold 0.6. By default, the ground truth heatmap is generated by GC with $\eta = \phi = 0.5$. Additional training details are discussed in \cref{app:imp}.

\vtb{Evaluation metrics.} We adopt the proposed CAS (\cref{sec:cas}) as the primary metric to evaluate model performance. Inspired by COCO’s AP, we also provide CAS evaluations across small ($\CAS_S$), medium ($\CAS_M$), and large ($\CAS_L$) bounding box types. We include precision (P), recall (R), and F1 scores in the discussion in \cref{sec:sota} to demonstrate their correspondence to MD, CP, and CAS, respectively. To quantify P and R for object center detection, only matched predicted points are considered as positives, and they are further regarded as true positives only if they fall within the corresponding ground truth bounding box.

%-------------------------------------------------------------------------
\subsection{Comparison with State-of-the-Art Frameworks}
\label{sec:sota}

We compare the proposed OCDet framework with state-of-the-art object detection and keypoint detection frameworks for the 80-class object center detection task on the COCO dataset. The main results are presented in \cref{fig:cas_latency} and \cref{tab:main}.

We evaluate the proposed OCDet framework against leading real-time object detection models: YOLOv8 and YOLO11. The predicted object centers of YOLO models are extracted from their bounding box outputs. Our OCDet  demonstrates superior performance. Across all five model variants (N/S/M/L/X), our OCDet outperforms YOLO11 with significant improvements of 23\% / 1\% / 6\% / 11\% / 8\% in CAS, while requiring 42\% / 83\% / 61\% / 25\% / 61\% fewer parameters, 34\% / 65\% / 69\% / 30\% / 71\% less computation, and achieving 64\% / 54\% / 39\% / 28\% / 22\% lower latency. The performance gains extend to recall (R) and F1 scores, where OCDet achieves enhancements of up to 245\% and 140\%, respectively. When compared with YOLOv8, our models exhibit a similar trend of dominance. 

To compare with the standard keypoint detection framework, we construct a series of KP models by utilizing the same model structures and configurations as our OCDet models. Following the standard keypoint detection practice, KP's ground truth heatmaps are built by applying a 2D Gaussian with a fixed standard deviation ($\sigma=2$ as in \cite{wangDeepHighResolutionRepresentation2020}) at each center. OCDet, with its GC heatmap, outperforms the KP framework under identical model configurations, achieving enhancements of 186\% / 66\% / 35\% / 29\% / 25\% in CAS, 138\% / 85\% / 72\% / 59\% / 93\% in recall, and 63\% / 33\% / 31\% / 29\% / 47\% in F1 score.

Despite OCDet's overall superiority, keypoint and object detection frameworks offer valuable insights. YOLO11 excels in Matched Discrepancy (MD), while KP models achieve higher precision (P), suggesting that regression-based center detection like YOLO and a more concentrated probability heatmap as in KP could enhance the accuracy of detected object centers. However, this comes at the cost of significant increases in missing target objects, as reflected in their dramatically higher CP and lower R. Additionally, while OCDet dominates other frameworks in terms of $\CAS_M$ and $\CAS_L$, it falls slightly behind KP in $\CAS_S$. We attribute this to pixel-level quantization errors caused by small size, which directly diminishes alignment in the GC map. Nonetheless, GC heatmaps still provide more effective information than bounding boxes for small objects, as OCDet consistently surpasses both YOLOv8 and YOLO11 across all object sizes.

\begin{table*}[!htp]
\centering
\caption{Comparison of our proposed OCDet framework with state-of-the-art real-time object detectors YOLOv8, YOLO11, and the identical models trained under the standard Keypoint framework (KP) across five model variants. Model performance is assessed using the proposed Center Alignment Score (CAS) with its component Cardinality Penalty (CP) and Matched Discrepancy (MD). CAS is further evaluated across three box types ($\CAS_L$, $\CAS_M$, $\CAS_s$). The corresponding precision (P), recall (R), and F1 score are also provided.}
\label{tab:main}
\vspace{-1.5mm}
\resizebox{1.0\linewidth}{!}{
\renewcommand\arraystretch{1.2}
\setlength{\tabcolsep}{0.15cm}
\begin{tabular}{lcccccccccccc}\toprule
\textbf{Model} &\textbf{\#Params.(M)} &\textbf{FLOPs(G)} &\textbf{Latency(ms)} &\textbf{$\CAS\uparrow$} &\textbf{$\CP\downarrow$} &\textbf{$\MD\downarrow$} &\textbf{P}$\uparrow$ &\textbf{R}$\uparrow$ &\textbf{F1}$\uparrow$ &\textbf{$\CAS_L\uparrow$} &\textbf{$\CAS_M\uparrow$} &\textbf{$\CAS_S\uparrow$} \\\midrule
YOLOv8n &3.16 &1.10 &26.95 &0.240 &0.735 &\textbf{0.025} &0.752 &0.195 &0.292 &0.397 &0.378 &0.096 \\
YOLO11n &2.62 &0.82 &30.77 &0.241 &0.735 &\textbf{0.025} &0.754 &0.198 &0.295 &0.401 &0.377 &0.097 \\
KP-N &\textbf{1.51} &\textbf{0.54} &\textbf{10.94} &0.104 &0.870 &0.026 &\textbf{0.861} &0.135 &0.212 &0.380 &0.351 &\textbf{0.248} \\
\rowcolor[gray]{0.92}
OCDet-N &\textbf{1.51} &\textbf{0.54} &\textbf{10.94} &\textbf{0.297} &\textbf{0.645 }&0.059 &0.634 &\textbf{0.466} &\textbf{0.510} &\textbf{0.875} &\textbf{0.614} &0.159 \\\midrule
YOLOv8s &11.17 &3.59 &41.44 &0.312 &0.657 &0.031 &0.747 &0.281 &0.391 &0.443 &0.495 &0.169 \\
YOLO11s &9.46 &2.70 &52.23 &0.309 &0.662 &\textbf{0.030} &0.772 &0.273 &0.385 &0.440 &0.481 &0.165 \\
KP-S &\textbf{1.59} &\textbf{0.94} &\textbf{24.25} &0.188 &0.769 &0.043 &\textbf{0.870} &0.213 &0.326 &0.629 &0.419 &\textbf{0.298} \\
\rowcolor[gray]{0.92}
OCDet-S &\textbf{1.59} &\textbf{0.94} &\textbf{24.25} &\textbf{0.313} &\textbf{0.644} &0.055 &0.621 &\textbf{0.507} &\textbf{0.531} &\textbf{0.889} &\textbf{0.652} &0.193 \\\midrule
YOLOv8m &25.90 &9.89 &75.19 &0.357 &0.610 &0.033 &0.777 &0.324 &0.440 &0.464 &0.547 &0.227 \\
YOLO11m &20.11 &8.56 &103.26 &0.343 &0.628 &\textbf{0.029} &0.800 &0.305 &0.423 &0.456 &0.525 &0.202 \\
KP-M &\textbf{8.25} &\textbf{3.54} &\textbf{63.26} &0.269 &0.679 &0.052 &\textbf{0.898} &0.316 &0.450 &0.692 &0.507 &\textbf{0.330} \\
\rowcolor[gray]{0.92}
OCDet-M &\textbf{8.25} &\textbf{3.54} &\textbf{63.26} &\textbf{0.362} &\textbf{0.573} &0.065 &0.643 &\textbf{0.585} &\textbf{0.599} &\textbf{0.895} &\textbf{0.718} &0.282 \\\midrule
YOLOv8l &43.69 &20.68 &125.88 &0.368 &0.600 &0.032 &0.804 &0.322 &0.442 &0.463 &0.549 &0.241 \\
YOLO11l &25.37 &10.92 &131.51 &0.349 &0.623 &\textbf{0.028} &0.808 &0.311 &0.431 &0.459 &0.527 &0.211 \\
KP-L &\textbf{31.88} &\textbf{7.67} &\textbf{94.14} &0.302 &0.642 &0.056 &\textbf{0.870} &0.347 &0.480 &0.507 &0.557 &\textbf{0.365} \\
\rowcolor[gray]{0.92}
OCDet-L &\textbf{31.88} &\textbf{7.67} &\textbf{94.14} &\textbf{0.389} &\textbf{0.563} &0.064 &0.694 &\textbf{0.596} &\textbf{0.630} &\textbf{0.894} &\textbf{0.724} &0.313 \\\midrule
YOLOv8x &68.23 &32.27 &\textbf{164.02} &0.376 &0.591 &0.033 &0.812 &0.333 &0.454 &0.471 &0.555 &0.255 \\
YOLO11x &56.97 &24.44 &206.34 &0.378 &0.590 &\textbf{0.032} &0.814 &0.337 &0.460 &0.460 &0.557 &0.256 \\
KP-X &\textbf{22.20} &\textbf{7.00} &181.81 &0.327 &0.617 &0.056 &\textbf{0.830} &0.405 &0.517 &0.330 &0.583 &\textbf{0.442} \\
\rowcolor[gray]{0.92}
OCDet-X &\textbf{22.20} &\textbf{7.00} &181.81 &\textbf{0.410} &\textbf{0.525} &0.066 &0.713 &\textbf{0.643} &\textbf{0.669} &\textbf{0.911} &\textbf{0.761} &0.324 \\
\bottomrule
\end{tabular}
}
\vspace{-2mm}
\end{table*}

\begin{table}[!htp]
\vspace{-2mm}
\centering
\caption{Ablation study on ground truth types.}
\label{tab:gt}
\vspace{-2mm}
\begin{tabular}{lrrrr}\toprule
\textbf{Ground truth type} &\textbf{$\CAS\uparrow$} &\textbf{$\CP\downarrow$} &\textbf{$\MD\downarrow$} \\\midrule
Gaussian (KP) &0.378 &0.550 &\textbf{0.071} \\
Ellipse &0.493 &0.421 &0.086 \\
\rowcolor[gray]{0.92}
GC (Ours) &\textbf{0.506} &\textbf{0.411} &0.083 \\\midrule
w/ segmentation &0.474 &0.446 &0.079 \\
w/o deep supervision &0.503 &0.420 &0.075 \\
\bottomrule
\end{tabular}
% \vspace{-2mm}
\end{table}

\begin{table}[!htp]
\centering
\caption{Ablation study on $\eta$ and $\phi$ for GC.}
\label{tab:eta_phi}
\vspace{-1.5mm}
\begin{tabular}{ccccc}\toprule
$\eta$ &$\phi$ &$\CAS_{\person}\uparrow$ &$\CAS_{\skis}\uparrow$ &$\CAS_{\bottle}\uparrow$ \\\midrule
0.1 &0.1 &0.447 &0.374 &0.167 \\
0.3 &0.3 &0.486 &0.446 &0.203 \\
0.5 &0.5 &\textbf{0.506} &0.457 &0.278 \\
0.7 &0.7 &\textbf{0.506} &0.426 &0.267 \\
0.9 &0.9 &0.475 &0.404 &0.275 \\\midrule
0.1 &0.9 &0.477 &0.376 &\textbf{0.283} \\
0.3 &0.7 &0.500 &0.412 &0.279 \\
0.7 &0.3 &0.493 &0.464 &0.210 \\
0.9 &0.1 &0.387 &\textbf{0.490} &0.147 \\
\bottomrule
\end{tabular}
\vspace{-2mm}
\end{table}

\begin{table}[!htp]
\vspace{-2mm}
\centering
\caption{Comparisons of proposed BCFL with other losses.}
\label{tab:loss}
\vspace{-2mm}
\begin{tabular}{lcccc}\toprule
\textbf{Loss} &\textbf{CAS} &\textbf{$\CP\downarrow$} &\textbf{$\MD\downarrow$} &\textbf{$\CAS_S\uparrow$} \\\midrule
BCE &0.312 &0.650 &0.038 &0.016 \\
MSE &0.337 &0.629 &0.034 &0.018 \\
QFL &0.364 &0.603 &\textbf{0.033 } &0.044\\\midrule
Weighted BCE &0.384 &0.533 &0.083 &0.215 \\
Weighted MSE &0.430 &0.492 &0.078 &0.176 \\
\rowcolor[gray]{0.92}
BCFL &\textbf{0.506} &\textbf{0.411} &0.083 &\textbf{0.314} \\
\bottomrule
\end{tabular}
\vspace{-1mm}
\end{table}

\begin{table}[!htp]
\centering
\caption{Ablation study on $\alpha$ for BCFL.}
\label{tab:alpha}
\vspace{-2mm}
\begin{tabular}{ccccc}\toprule
\textbf{Threshold} &\textbf{$\alpha$} &\textbf{$\CAS\uparrow$} &\textbf{$\CP\downarrow$} &\textbf{$\MD\downarrow$} \\\midrule
- &0.500 &0.364 &0.603 &\textbf{0.033}\\
0 &0.723 &0.440 &0.507 &0.053 \\
0.2 &0.811 &0.450 &0.494 &0.056 \\
0.3 &0.860 &0.476 &0.463 &0.062 \\
0.4 &0.904 &0.496 &0.432 &0.072 \\
0.5 &0.938 &0.499 &0.427 &0.075 \\
\rowcolor[gray]{0.92}
0.6 &0.964 &\textbf{0.506} &\textbf{0.411} &0.083 \\
0.7 &0.981 &0.479 &0.435 &0.086 \\
0.8 &0.993 &0.403 &0.517 &0.080 \\
\bottomrule
\end{tabular}
\vspace{-2mm}
\end{table}

\begin{table}[!htp]
\centering
\caption{Ablation study on $\gamma$ for BCFL.}
\label{tab:gamma}
\vspace{-1.5mm}
\begin{tabular}{cccccc}\toprule
\textbf{$\gamma$} &\textbf{$\CAS\uparrow$} &\textbf{$\CP\downarrow$} &\textbf{$\MD\downarrow$} &\textbf{$\CAS_M\uparrow$} &\textbf{$\CAS_S\uparrow$} \\\midrule
1 &0.473 &0.441 &0.085 &0.725 &0.300\\
\rowcolor[gray]{0.92}
2 &\textbf{0.506} &\textbf{0.411} &0.083 &\textbf{0.754} &\textbf{0.314} \\
3 &0.477 &0.441 &0.083 &0.683 &0.247 \\
4 &0.469 &0.454 &\textbf{0.077} &0.668 &0.216 \\
\bottomrule
\end{tabular}
\vspace{-3mm}
\end{table}

%-------------------------------------------------------------------------
\subsection{Ablation Studies}
\label{sec:ablation}

In this section, we present ablation studies on our proposed GC and BCFL. Unless otherwise specified, we use the OCDet-S model for center detection across all person instances in the COCO dataset. Additional ablation studies regarding model configurations, including backbone, architecture, module design, and postprocessing, are discussed in \cref{app:ablation}.

\vtb{Ground truth heatmap type.} \cref{tab:gt} summarizes the impact of ground truth heatmap formats on object center detection. In addition to fixed Gaussian as in the keypoint framework (KP), we adopt an ellipse-based approach inspired by \cite{dolezelCentroidBasedPerson2022}, assigning probabilities within the bounding box-inscribed ellipse according to the ellipse equation. Our results indicate that our GC outperforms the ellipse-based method by 0.013 in CAS, while both surpassing the conventional KP by more than 30\%. This highlights the benefit of bounding box-aware heatmaps for object center detection tasks. However, further incorporating the segmentation mask by cutting out the precise shape of the object from the original ground truth GC map reduces CAS by 7\% (row 4 of \cref{tab:gt}), likely due to the disruption of GC's symmetrical structure. Moreover, injecting ground truth information at multiple stages provides a slight advantage, as removing the deep supervision (detailed in \cref{app:imp}) leads to a slight degradation in all metrics.

\vtb{Generalized Centerness.} By adjusting the parameters $\eta$ and $\phi$ in \cref{eq:gc}, we can generate GC maps with varying shapes, as illustrated in \cref{fig:gc} in the supplementary material. \cref{tab:eta_phi} summarizes the model’s performance under various $\eta$ and $\phi$ values for center detection of three representative objects: person (regular), skis (wide), and bottle (elongated). Overall, isotropic ($\eta=\phi$) GC heatmaps with $\eta$ and $\phi$ around 0.5 and 0.7 demonstrate strong performance. These configurations achieve a balanced trade-off between CP and MD, resulting in high CAS in general. For elongated or wide objects, a GC map that exhibits the opposite shape of its own tends to achieve better performance. For instance, the most horizontally extended GC configuration with $\eta=0.1$ and $\phi=0.9$ yields the best result for tall objects like bottles, whereas for wide objects such as skis, the best performance is achieved with the most vertically spread GC setting of $\eta=0.9$ and $\phi=0.1$. When a horizontally extended GC map is applied to a wide object, the horizontal centerline is not only long due to the object's shape but also exhibits uniformly high values due to the GC configuration. It becomes more challenging for the model to localize the center. In contrast, an elongated GC heatmap assigns significantly higher values along the longitudinal centerline, allowing the model, when combined with BCFL, to rapidly identify the vertical center position. Furthermore, the rapid decay in the horizontal direction facilitates precise horizontal center localization, which is particularly beneficial for wide objects. In summary, our proposed GC can dynamically adapt to different object shapes, enabling more accurate center point localization.

\vtb{Balanced Continuous Focal Loss.} Results using various losses are listed in \cref{tab:loss}. We divide the loss functions into two groups based on whether they incorporate specific handling for class imbalance. In the first group in \cref{tab:loss} without imbalance handling, QFL shows a CAS improvement of 0.052 and 0.027 compared to commonly used BCE and MSE, demonstrating the importance of hard negatives for continuous probability regression tasks. In the second group, we apply fixed weighting to BCE and MSE as well as our proposed $\ac$-balancing to BCFL (\cref{eq:bcfl}). Weighted BCE and MSE bring significant CAS improvements of 23\% and 27\%, demonstrating the critical role of class imbalance handling in object center detection. More importantly, incorporating our $\ac$-balancing with the target-dependent weighting mechanism boosts overall CAS by nearly 40\% and $\CAS_S$ for small objects by over 7 times, further affirming the effectiveness of our proposed BCFL.

We further investigate the best hyperparameter setups for BCFL. As listed in \cref{tab:gamma}, $\gamma=2$ (in \cref{eq:bcfl}) achieves optimal performance across all metrics, consistent with the findings in \cite{linFocalLossDense2018,liGeneralizedFocalLoss2020a}. We calculate $\alpha$ based on the inverse class frequency given a probability threshold. In \cref{tab:alpha}, we determine that the optimal probability threshold is 0.6, which corresponds to $\alpha=0.964$ for person center detection. The probability threshold of 0.6 is uniformly applied across all experiments for the calculation of the optimal $\alpha$.

%% file: sec/5_conclusion.tex
\section{Conclusion}
\label{sec:conclusion}

In this paper, we present OCDet, a novel object center detection framework tailored to real-world applications that require real-time, coarse object localization with limited computational resources. By employing bounding box-aware heatmap generation with Generalized Centerness (GC), OCDet can accurately identify object centers without the need for extra manual labeling. The proposed Balanced Continuous Focal Loss (BCFL) addresses data imbalance and directs training attention to hard negative examples, making it particularly suitable for probability regression tasks with skewed target distributions like GC heatmap-based object center detection. We define a novel Center Alignment Score (CAS) metric to incorporate the concept of precision and recall into a normalized distance metric with the help of Hungarian matching. Extensive experiments demonstrate that OCDet outperforms state-of-the-art object detectors and standard keypoint detection frameworks by a large margin, achieving significantly higher CAS, recall, and F1 scores with substantially reduced latency, fewer parameters, and lower FLOPs, These results establish OCDet as an efficient and robust solution for real-time object localization on edge devices with NPUs.

%% file: sec/6_suppl.tex
\clearpage
\setcounter{page}{1}
\maketitlesupplementary

\appendix

%-------------------------------------------------------------------------
\section{Figures}
\label{app:figures}

\begin{figure*}[ht]
  \centering
   \includegraphics[width=1.0\linewidth]{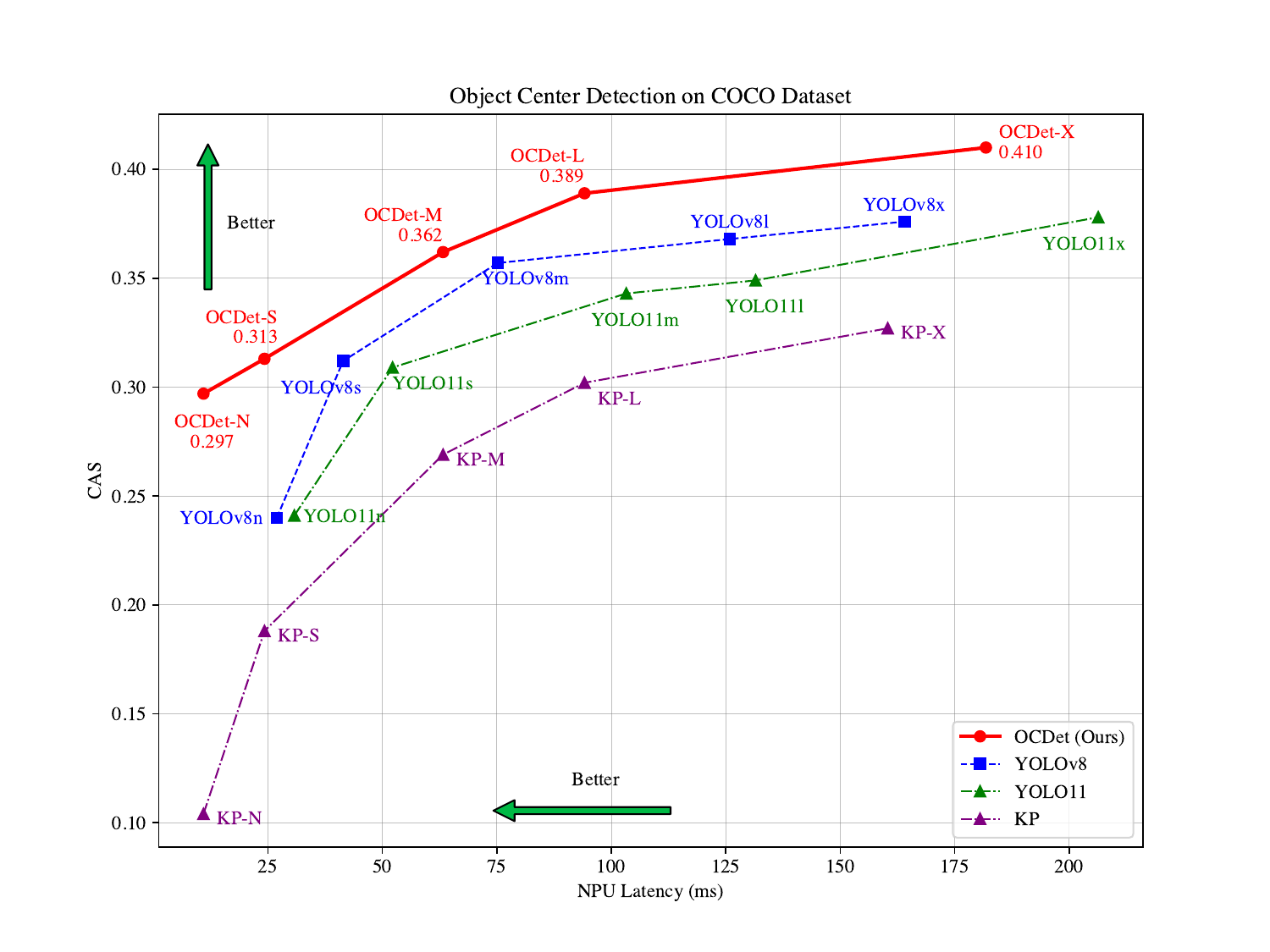}
   \vspace{-12mm}
   \caption{Evaluation of the proposed OCDet against state-of-the-art real-time object detectors YOLOv8, YOLO11, and the same models trained under a standard keypoint-based framework (KP). Corresponds to \cref{fig:cas_latency} in the main paper.}
   \label{fig:cas_latency_details}
\end{figure*}

\cref{fig:cas_latency_details} enlarges \cref{fig:cas_latency} from the main paper, providing a detailed view of each model variant.

\section{Visualization of Generalized Centerness heatmaps}
\label{app:gc}
\begin{figure}[ht]
    \centering
    \includegraphics[width=1\linewidth]{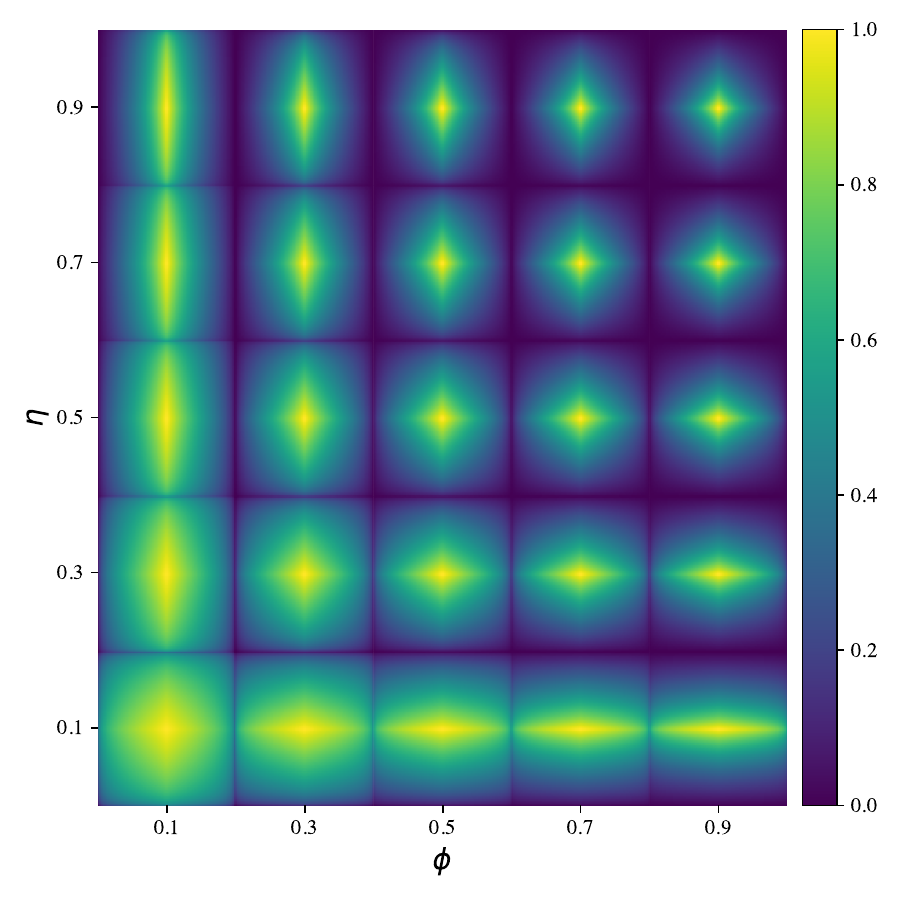}
    \vspace{-0.8cm}
    \caption{Illustration of Generalized Centerness heatmaps of a square bounding box under different $\eta$ and $\phi$ configurations.}
    \label{fig:gc}
\end{figure}

\cref{fig:gc} demonstrates the shape of the Generalized Centerness (GC) as it varies with the controlling hyperparameters $\eta$ and $\phi$ in \cref{eq:gc}. Increasing $\eta$ and $\phi$ results in a more concentrated heatmap around the center, whereas decreasing $\eta$ and $\phi$ disperses the concentration along the horizontal and vertical axes, respectively. When $\eta$ is large and $\phi$ is small (top-left region of \cref{fig:gc}), the GC map exhibits an elongated shape with high values concentrated along the vertical axis and quickly fades along the horizontal direction. Conversely, a small $\eta$ combined with large $\phi$ (bottom-right region of \cref{fig:gc}) results in a horizontally spread distribution.

%-------------------------------------------------------------------------
\section{More Model Details}
\label{app:models}
\ntb{Backbone.} Due to the constraints on NPU, we choose the MobileNetV4 family \cite{qinMobileNetV4UniversalModels2024} as the backbone of the proposed OCDet models. The MobileNetV4 backbone mainly consists of three novel blocks: Universal Inverted Bottleneck (UIB), Inverted Bottleneck (IB), and Extra Depthwise (ExtraDW). The blocks utilize a sequential combination of depthwise, pointwise, and standard convolutional layers, interspersed with canonical ReLU and BatchNorm. By capitalizing on these well-optimized operations, OCDet models achieve exceptional computational efficiency on NPUs.

\begin{figure*}[ht]
    \centering
    \includegraphics[width=1.0\linewidth]{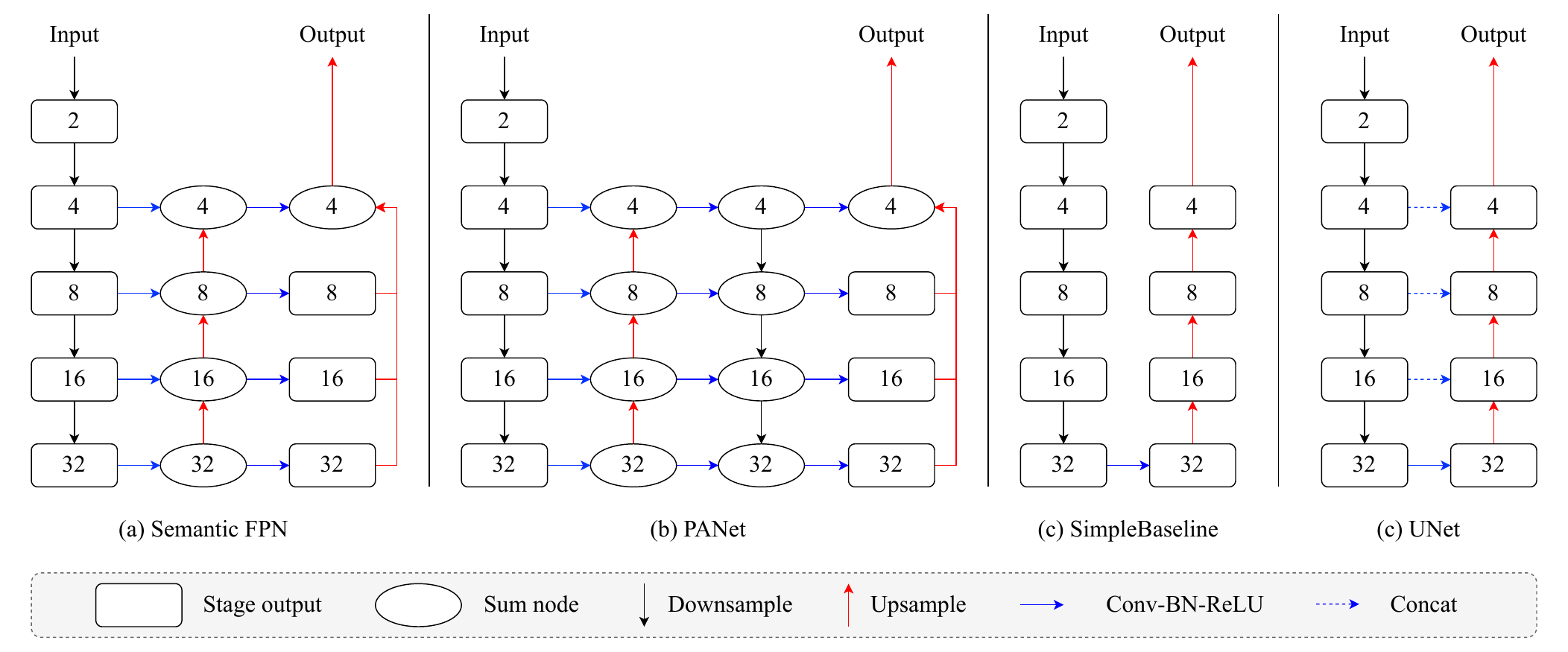}
    \vspace{-5mm}
    \caption{Architecture diagrams of (a) Semantic FPN \cite{kirillovPanopticFeaturePyramid2019}, PANet \cite{liuPathAggregationNetwork2018}, UNet \cite{ronnebergerUNetConvolutionalNetworks2015}, and SimpleBaseline \cite{xiaoSimpleBaselinesHuman2018}}
    \label{fig:models}
\end{figure*}

\vtb{Architecture.} OCDet models adopt the Semantic FPN architecture \cite{kirillovPanopticFeaturePyramid2019}. As shown in the figure in \cref{fig:models} (a), FPN \cite{linFeaturePyramidNetworks2017} fuses features from the final stages of MobileNetV4 (left column). This feature fusion utilizes standard Conv-BN-RelU and bilinear upsampling modules to construct a multi-scale feature pyramid (middle column). OCDet models leverage the last three or four levels of the pyramid (see \cref{app:ablation}), corresponding to feature maps with resolutions ranging from 1/8 (for three levels) or 1/4 (for four levels) to 1/32 of the input resolution. Finally, the Semantic FPN architecture upsamples these augmented feature maps to the target output resolution and sums them up to produce the final model output (right column). This simple structure of FPN demonstrates impressive speed performance on NPUs.
 
\vtb{Alternative architectures.} In addition to the employed Semantic FPN, we evaluate the performance of other NPU-friendly architectures such as UNet \cite{ronnebergerUNetConvolutionalNetworks2015}, SimpleBaseline \cite{xiaoSimpleBaselinesHuman2018}, and PANet \cite{liuPathAggregationNetwork2018} across S/M/L backbone configurations. \cref{fig:models} presents diagrams of these architectures. PANet (\cref{fig:models} (b)), building upon FPN, incorporates an extra bottom-up path augmentation path to strengthen feature fusions. SimpleBaseline (\cref{fig:models} (c)) is a lightweight single-path model with a minimalistic design. It employs a standard backbone for feature extraction and integrates several deconvolutional layers to generate high-resolution heatmaps. U-Net (\cref{fig:models} (d)) employs an encoder-decoder architecture with skip connections that concatenate features from the encoder and decoder to fuse low-level spatial details with high-level contextual information.

We adapt the original implementations of these models to suit the object center detection task. First, all backbones are replaced with MobileNetV4 models. For upsampling, we employ bilinear interpolation followed by a Conv-BN-ReLU module across all models. In addition,deconvolution-based upsampling is implemented for UNet and SimpleBaseline. To achieve an output stride of 4, corresponding to an output resolution that is one-quarter of the input resolution, we attach the same Semantic FPN head to the feature pyramid extracted by PANet. In the case of SimpleBaseline, we implement a four-stage upsampling process scaling from 1/32 to 1/4. For UNet, the final two stages and their associated skip connections in the upsampling branch are removed. Consequently, the feature map from the last remaining stage, already at the target output stride of 4, is directly used as the output. Besides the output stride of 4, PANet and FPN also include versions with an output stride of 8. In these cases, only the last three levels of the feature pyramid are fused, as opposed to the last four levels used for the output stride of 4.

%-------------------------------------------------------------------------
\section{More Implementation Details}
\label{app:imp}

\ntb{Model configurations.} Based on MobileNetV4 and Semantic FPN, we build a series of OCDet models with five variants: N/S/M/L/X. OCDet-S/M/L uses MobileNetV4’s convolutional variants, MobileNetV4-Conv-S/M/L, as their respective backbones. Inspired by the scaling mechanism of YOLOv8 \cite{jocherUltralyticsYOLOv82023} and YOLO11 \cite{jocherUltralyticsYOLO112024}, we further introduce OCDet-N and OCDet-X at either end of the OCDet model family spectrum. The former is a truncated version of OCDet-S, utilizing only the final three levels of the feature pyramid, yielding an output stride of 8. The latter retains the same four-level FPN architecture with an output stride of 4 but employs the more powerful EfficientNetV2-S \cite{tanEfficientNetV2SmallerModels2021} as the backbone.

\vtb{Training Details.} Model training is performed on an NVIDIA A100 GPU with float32. We initialize the model backbone with the official ImageNet-pre-trained weights, while the remaining convolutional layers are initialized using Kaiming initialization\cite{heDelvingDeepRectifiers2015}. All models are trained for 24 epochs using the 2× schedule with a batch size of 64. We adopt AdamW \cite{loshchilovDecoupledWeightDecay2017} optimizer with a weight decay of $3 \times 10^{-4}$. We set $\beta_1=0.931$ and $\beta_2=0.997$. The initial learning rate is adjusted based on model size, ranging from $7 \times 10^{-4}$ to $6 \times 10^{-3}$. After a warm-up period of up to 1200 steps, the learning rate gradually decays to zero according to a cosine schedule. The loss function is BCFL (\cref{eq:bcfl}) with $\gamma = 2$, and $\alpha$ set to 0.964 and 0.984 for person and 80-class object center detection, respectively. The $\alpha$ values correspond to the inverse class frequencies calculated based on the probability threshold of 0.6. The ground truth heatmap is generated by \cref{eq:gc} with $\eta=\phi=0.5$ by default. Slightly modified from previous heatmap-based keypoint detection frameworks \cite{wangDeepHighResolutionRepresentation2020, newellAssociativeEmbeddingEndEnd2017, xiaoSimpleBaselinesHuman2018}, our data augmentation includes random rotations within $[-30\degree, 30\degree]$, scaling within $[0.75, 1.25]$, as well as random horizontal and vertical flips. The input image is resized to $320\times320$ for both training and inference. We employ deep supervision by supervising both the upsampled model output with resized ground truth heatmaps at $320\times320$ and the original model output with ground truth heatmaps generated directly at $80\times80$ or $40\times40$, depending on the model configuration.

\begin{table*}[!htp]
\centering
\caption{Comparisons of backbones. The tested backbones are categorized into four groups based on latency. Within each group, the MobileNetV4 model, adopted by our OCDet framework, consistently demonstrates superior performance, achieving higher CAS with lower latency.}
\label{tab:backbone}
\begin{tabular}{llcccccc}\toprule
\textbf{Architecture} & \textbf{Backbone} &\textbf{\#Params.(M)} &\textbf{FLOPs(G)} &\textbf{Latency(ms)} &\textbf{$\CAS\uparrow$} &\textbf{CP $\downarrow$} \\\midrule
\rowcolor[gray]{0.92}
FPN (OCDet-S)  & MobileNetV4-Conv-S &1.59 &0.92 &\textbf{18.41} &\textbf{0.506} &\textbf{0.411 }\\
FPN & MobileNetV2-$\times$0.75 &1.60 &0.89 &26.01 &0.444 &0.475 \\
FPN & MobileNetV3-S &1.16 &0.54 &28.21 &0.464 &0.453 \\\midrule
\rowcolor[gray]{0.92}
FPN (OCDet-M) & MobileNetV4-Conv-M &7.79 &2.75 &\textbf{39.68} &\textbf{0.546} &0.379 \\
FPN & MobileNetV2-$\times$1.4 &4.31 &2.71 &43.83 &0.542 &\textbf{0.376} \\
FPN & MobileNetV3-L &3.79 &1.92 &51.53 &0.528 &0.387 \\\midrule
\rowcolor[gray]{0.92}
FPN (OCDet-L) & MobileNetV4-Conv-L &30.96 &6.06 &\textbf{60.47} &\textbf{0.557 }&\textbf{0.366 }\\
FPN & RepVGG-A2 &28.07 &13.81 &62.66 &0.542 &0.381 \\
FPN & MobileOne-S0 &4.82 &3.15 &77.61 &0.509 &0.405 \\\midrule
\rowcolor[gray]{0.92}
FPN (OCDet-X) & EfficientNetV2-S &21.91 &6.49 &\textbf{160.37} &\textbf{0.582} &\textbf{0.337 }\\
FPN & EfficientNet-B3 &10.59 &2.86 &161.00&0.578 &0.339 \\
FPN & HRNet-W18 &12.19 &9.40 &172.61 &0.564 &0.357 \\
\bottomrule
\end{tabular}
\end{table*}

\begin{table*}[t]
\centering
\caption{Comparison of architectures. Using the same backbone, the FPN with an output stride of 8 outperforms UNet, SimpleBaseline, and their corresponding PANet variants, achieving a higher CAS while maintaining comparable or lower latency. Moreover, the FPN with an output stride of 4 always achieves the highest CAS among all configurations.}
\label{tab:arch}
\begin{tabular}{lclccccc}\toprule
\textbf{Architecture} &\textbf{Mode} &\textbf{Backbone} &\textbf{\#Params.(M)} &\textbf{FLOPs(G)} &\textbf{Latency(ms)} &\textbf{$\CAS\uparrow$} &\textbf{CP $\downarrow$} \\\midrule
UNet &bilinear &MobileNetV4-Conv-S &1.89 &0.86 &9.16 &0.454 &0.477 \\
UNet &deconv &MobileNetV4-Conv-S &2.28 &1.00 &9.21 &0.434 &0.484 \\
SimpleBaseline &bilinear &MobileNetV4-Conv-S &3.12 &0.72 &\textbf{8.98} &0.455 &0.466 \\
SimpleBaseline &deconv &MobileNetV4-Conv-S &3.91 &0.61 &9.18 &0.434 &0.494 \\
PANet &stride 8 &MobileNetV4-Conv-S &1.66 &0.56 &10.18 &0.469 &0.447 \\
\rowcolor[gray]{0.92}
FPN (OCDet-N) &stride 8 &MobileNetV4-Conv-S &1.51 &0.52 &9.99 &\textbf{0.472} &\textbf{0.440} \\\greyrule
PANet &stride 4 &MobileNetV4-Conv-S &1.81 &1.08 &18.80 &0.497 &0.429 \\
\rowcolor[gray]{0.92}
FPN (OCDet-S) &stride 4 &MobileNetV4-Conv-S &1.59 &0.92 &\textbf{18.41} &\textbf{0.506} &\textbf{0.411 }\\\midrule
UNet &bilinear &MobileNetV4-Conv-M &9.44 &3.56 &30.84 &0.477 &0.452 \\
UNet &deconv &MobileNetV4-Conv-M &10.76 &4.09 &31.53 &0.472 &0.449 \\
SimpleBaseline &bilinear &MobileNetV4-Conv-M &9.57 &2.45 &\textbf{29.20} &0.505 &0.425 \\
SimpleBaseline &deconv &MobileNetV4-Conv-M &12.08 &2.39 &30.17 &0.483 &0.443 \\
PANet &stride 8 &MobileNetV4-Conv-M &8.24 &2.21 &29.70 &0.495 &0.417 \\
\rowcolor[gray]{0.92}
FPN &stride 8 &MobileNetV4-Conv-M &7.66 &1.97 &29.41 &\textbf{0.510} &\textbf{0.405} \\\greyrule
PANet &stride 4 &MobileNetV4-Conv-M &8.29 &3.09 &40.22 &0.534 &0.382 \\
\rowcolor[gray]{0.92}
FPN (OCDet-M) &stride 4 &MobileNetV4-Conv-M &7.79 &2.75 &\textbf{39.68} &\textbf{0.546} &\textbf{0.379} \\\midrule
UNet &bilinear &MobileNetV4-Conv-L &34.90 &8.58 &55.43 &0.488 &0.433 \\
UNet &deconv &MobileNetV4-Conv-L &37.69 &9.76 &57.53 &0.479 &0.436 \\
SimpleBaseline &bilinear &MobileNetV4-Conv-L &32.74 &5.75 &51.89 &0.514 &0.416 \\
SimpleBaseline &deconv &MobileNetV4-Conv-L &40.20 &6.63 &55.02 &0.510 &0.418 \\
PANet &stride 8 &MobileNetV4-Conv-L &31.36 &5.02 &51.13 &0.509 &0.419 \\
\rowcolor[gray]{0.92}
FPN &stride 8 &MobileNetV4-Conv-L &30.77 &4.88 &\textbf{50.42} &\textbf{0.518} &\textbf{0.405} \\\greyrule
PANet &stride 4 &MobileNetV4-Conv-L &31.85 &6.68 &62.30 &0.541 &0.383 \\
\rowcolor[gray]{0.92}
FPN (OCDet-L) &stride 4 &MobileNetV4-Conv-L &30.96 &6.06 &\textbf{60.47} &\textbf{0.557} &\textbf{0.366} \\
\bottomrule
\end{tabular}
\end{table*}

%-------------------------------------------------------------------------
\section{Ablation Studies on Model Design}
\label{app:ablation}

In this section, we present ablation studies focused on the design of our OCDet models. As in \cref{sec:ablation}, we use the OCDet-S model for person center detection on the COCO dataset by default.

\vtb{Backbones.} In comparison with the adopted MobileNetV4 backbones, we evaluate the performance of promising convolutional backbones, such as MobileNetV2 \cite{sandlerMobileNetV2InvertedResiduals2018}, MobileNetV3 \cite{howardSearchingMobileNetV32019}, MobileOne \cite{vasuMobileOneImprovedOne2023}, HRNet \cite{wangDeepHighResolutionRepresentation2020, yuLiteHRNetLightweightHighResolution2021}, RepVGG \cite{dingRepVGGMakingVGGstyle2021}, EfficientNetV1 \cite{tanEfficientNetRethinkingModel2020}, and EfficientNetV2 \cite{tanEfficientNetV2SmallerModels2021}, on the object center detection task using the same Semantic FPN architecture with an output stride of 4. In \cref{tab:backbone}, the results are categorized into four groups based on latency. Across all groups, OCDet's backbones, especially the MobileNetV4 family, consistently outperform the comparison models, achieving higher CAS with lower latency. This superior performance is primarily due to the use of highly optimized standard convolutional blocks tailored for NPU efficiency. As a counterexample, SE blocks used in EfficientNet suffer from synchronization issues \cite{vasuMobileOneImprovedOne2023}. Consequently, despite relatively low FLOPs and parameter counts, its NPU latency remains high. EfficientNetV2's Fused-MBConv partially alleviates this issue, making it a suitable choice as the backbone of our largest OCDet-X.

\vtb{Architecture.} \cref{tab:arch} presents the statistics and evaluation results of OCDet's FPN architecture and three other architectures mentioned in \cref{app:models}. FPN stands out as the most effective architecture in the evaluation. Despite its more complex structure, PANet is consistently Pareto-dominated by FPN across all configurations and thus excluded from consideration. Compared to the best variant of UNet and SimpleBaseline across S/M/L backbones, FPN with the output stride of 8 consistently achieves higher CAS under similar latency, achieving CAS improvements of 0.021/0.005/0.004, respectively. Between the FPNs with output strides of 8 and 4, there's a consistent latency difference of around 10ms, solely attributed to the additional processing of the feature maps of 1/4 resolution from the backbone. Conversely, the CAS difference between them does not remain constant but progressively expands from 0.31 to 0.36 and eventually to 0.39, driven by richer high-level features provided by larger backbones. As a result, small models with an output stride of 8 demonstrate a better accuracy-latency tradeoff due to the consistent 10ms latency overhead introduced by the FPN. In contrast, for larger models, this 10ms difference becomes negligible compared to the dominant latency contributed by the backbone, making models with an output stride of 4 more advantageous. Accordingly, we default to an output stride of 4 for constructing OCDet-S/M/L/X and additionally introduce OCDet-N, which employs the same MobileNetV4-Conv-S backbone but with an output stride of 8.

\vtb{Output stride.} Given the superior performance of FPN, we further investigate its structural design. \cref{tab:num_outs} shows the model performance with output strides of 16/8/4/2, corresponding to the usage of the feature maps from the last 2/3/4/5 stages of the backbone, respectively. The results indicate that using only the feature maps from the last two stages (output stride 16)  degrades CAS by more than 10\%, whereas including all five stages (output stride 2) significantly increases latency by four times. Output strides 8 and 4, however, achieve a favorable balance between latency and CAS. Therefore, OCDet models primarily adopt the output stride of 4 for OCDet-S/M/L/X, while the smallest model, OCDet-N, is implemented by FPN with an output stride of 8.

\vtb{FPN module Design.} For the module design of FPN, we explore integrating novel components such as MobileNetV4 modules. Specifically, we evaluate the use of ConvNeXt, IB, and ExtraDW from MobileNetV4 as replacements for the traditional Conv-BN-ReLU module. As shown in \cref{tab:fpn_type}, while these alternatives have fewer parameters and FLOPs compared to the standard Conv-BN-ReLU module and exhibit similar CAS performance, the latency of the latter remains the lowest due to its extensive optimization on the NPU. Therefore, we retain the standard Conv-BN-ReLU module as the building block for FPN.

\begin{figure}[ht]
    \centering
    \begin{subfigure}{\linewidth}
        \centering
        \includegraphics[width=0.8\linewidth]{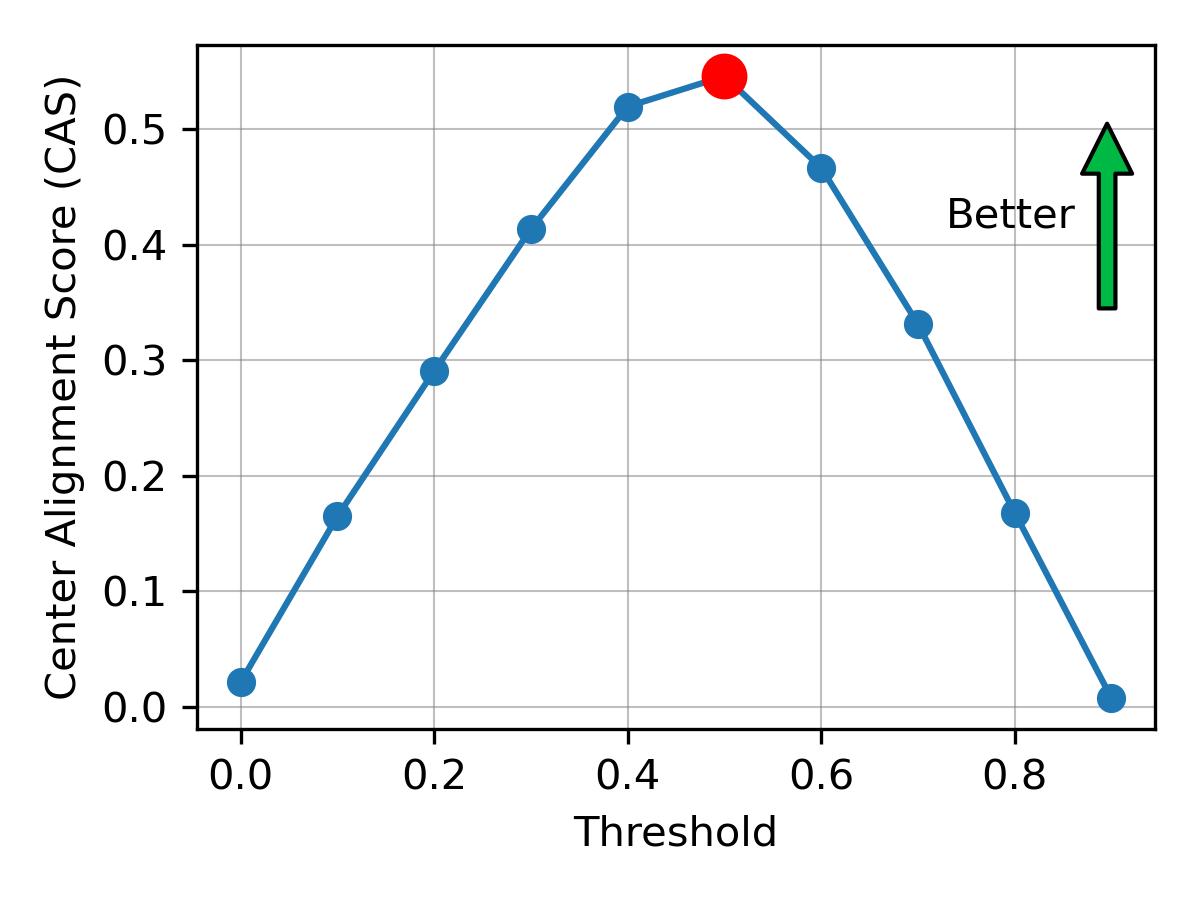}
        \vspace{-1.5mm}
        \caption{CAS across thresholds with a fixed minimum distance of 0.3}
        \vspace{2mm}
    \end{subfigure}
    \begin{subfigure}{\linewidth}
        \centering
        \includegraphics[width=0.8\linewidth]{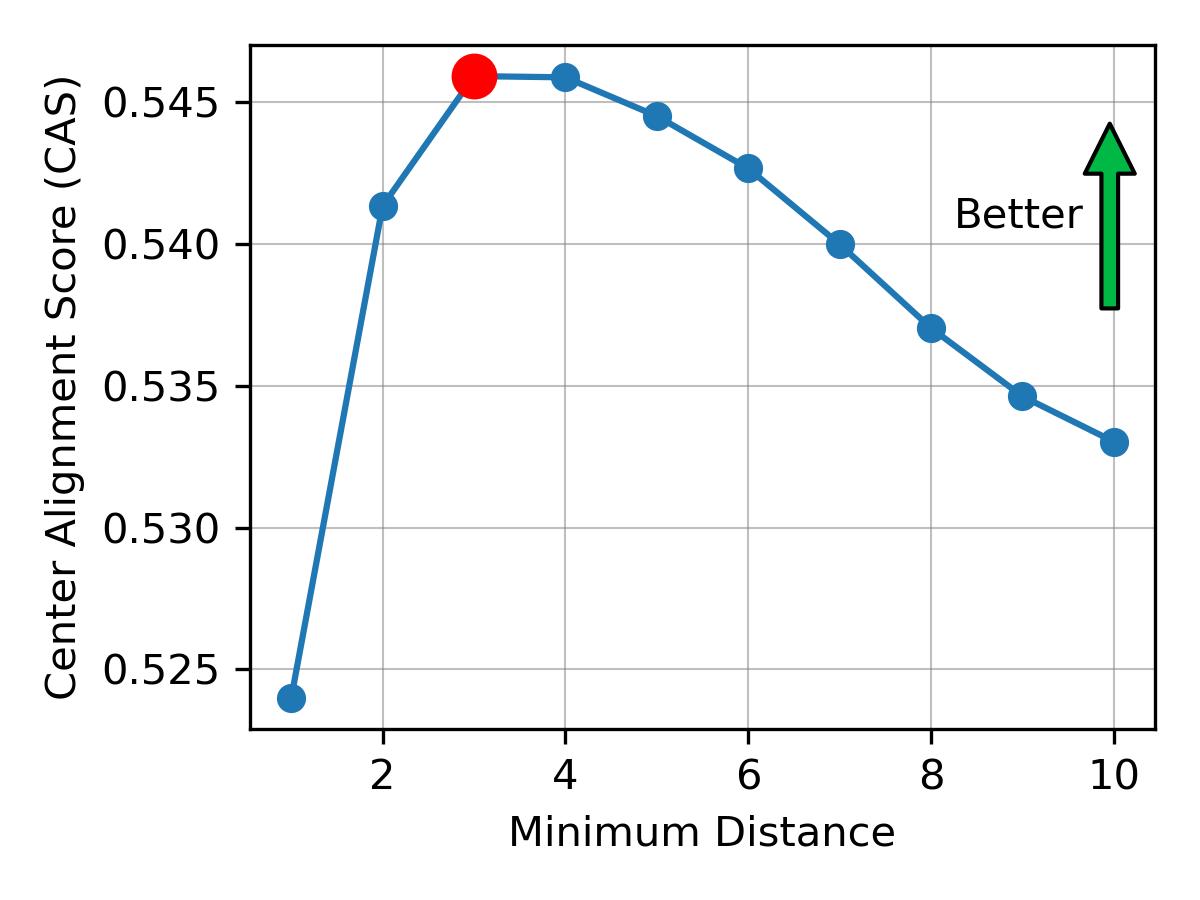}
        \vspace{-1.5mm}
        \caption{CAS across minimum distances with a fixed threshold of 0.5}
    \end{subfigure}
    \vspace{-5mm}
    \caption{Visualization of the ablation study for peak identification}
    \label{fig:ablation_postprocessing}
\end{figure}

\vtb{Peak Identification.} We investigate the impact of two critical hyperparameters in peak identification: the probability threshold and the minimum distance between predicted centers, leveraging the OCDet-M model. \cref{fig:ablation_postprocessing} (a) illustrates how CAS responds to varying probability thresholds under a fixed minimum distance. The results show that CAS exhibits optimal performance at a probability threshold of 0.5. Similarly, \cref{fig:ablation_postprocessing} (b) presents the changes in CAS with respect to the minimum distance, revealing that the optimal value is 0.3. These selected values are consistently used across all experiments.

\begin{table}[ht]
\centering
\caption{Ablation study on output stride.}
\label{tab:num_outs}
\vspace{-2mm}
\resizebox{1.0\linewidth}{!}{
\renewcommand\arraystretch{1.2}
\setlength{\tabcolsep}{0.15cm}
\begin{tabular}{ccccc}\toprule
\textbf{Output stride} &\textbf{\#Params.(M)} &\textbf{FLOPs(G)} &\textbf{Latency(ms)} &\textbf{$\CAS\uparrow$} \\\midrule
16  &1.44 &0.42 &\textbf{8.11} &0.426 \\
\rowcolor[gray]{0.92}
8  &1.51 &0.52 &9.99 &0.475 \\
\rowcolor[gray]{0.92}
4  &1.59 &0.92 &18.41 &\textbf{0.506 }\\
2  &1.67 &2.60 &63.80 &0.497 \\
\bottomrule
\end{tabular}
}
\end{table}

\begin{table}[ht]
\centering
\caption{Ablation study on FPN modules.}
\label{tab:fpn_type}
\vspace{-2mm}
\resizebox{1.0\linewidth}{!}{
\renewcommand\arraystretch{1.2}
\setlength{\tabcolsep}{0.15cm}
\begin{tabular}{ccccc}\toprule
\textbf{Module}  &\textbf{\#Params.(M)} &\textbf{FLOPs(G)} &\textbf{Latency(ms)} &\textbf{$\CAS\uparrow$} \\\midrule
ConvNext  &1.62 &0.86 &19.40 &0.502 \\
IB  &1.48 &0.71 &21.66 &0.497 \\
ExtraDW  &1.49 &0.72 &26.94 &\textbf{0.507} \\
\rowcolor[gray]{0.92}
Conv-BN-ReLU  &1.59 &0.92 &\textbf{18.41} &0.506 \\
\bottomrule
\end{tabular}
}
\end{table}

%-------------------------------------------------------------------------
\section{Qualitative results}
\label{app:vis}

\begin{figure*}[ht]
    \centering
    \includegraphics[width=0.92\linewidth]{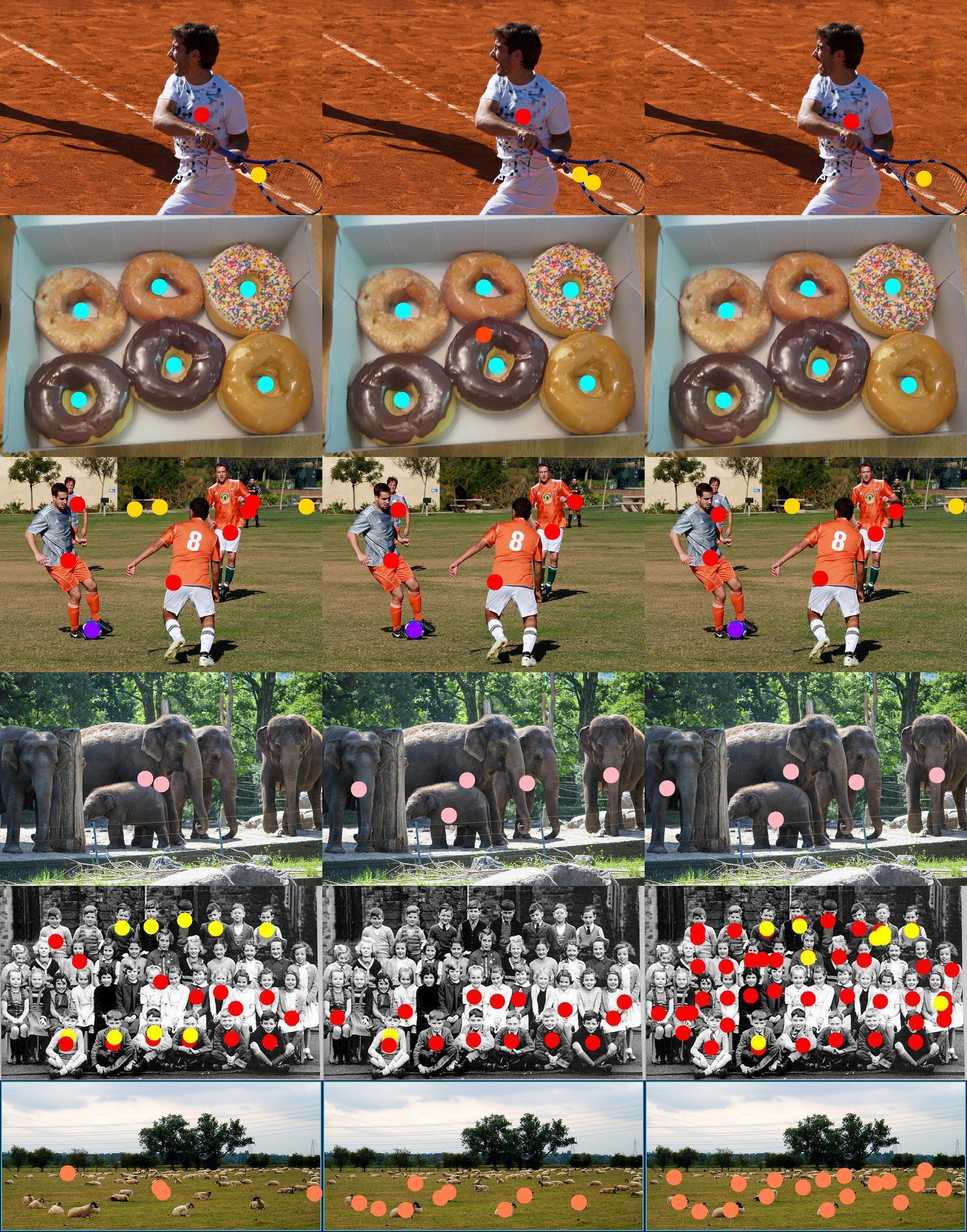}
    \begin{minipage}{0.33\linewidth}
        \centering
        Ground Truth
    \end{minipage}%
    \begin{minipage}{0.33\linewidth}
        \centering
        YOLO11x
    \end{minipage}%
    \begin{minipage}{0.33\linewidth}
        \centering
        OCDet-X (ours)
    \end{minipage}
    \caption{Comparison of OCDet-X (right), YOLO11x (center), and Ground Truth (left) for object center detection. Detected object centers are shown as colored dots.}
    \label{fig:comparison}
\end{figure*}

\begin{figure*}[ht]
    \centering
    \includegraphics[width=0.92\linewidth]{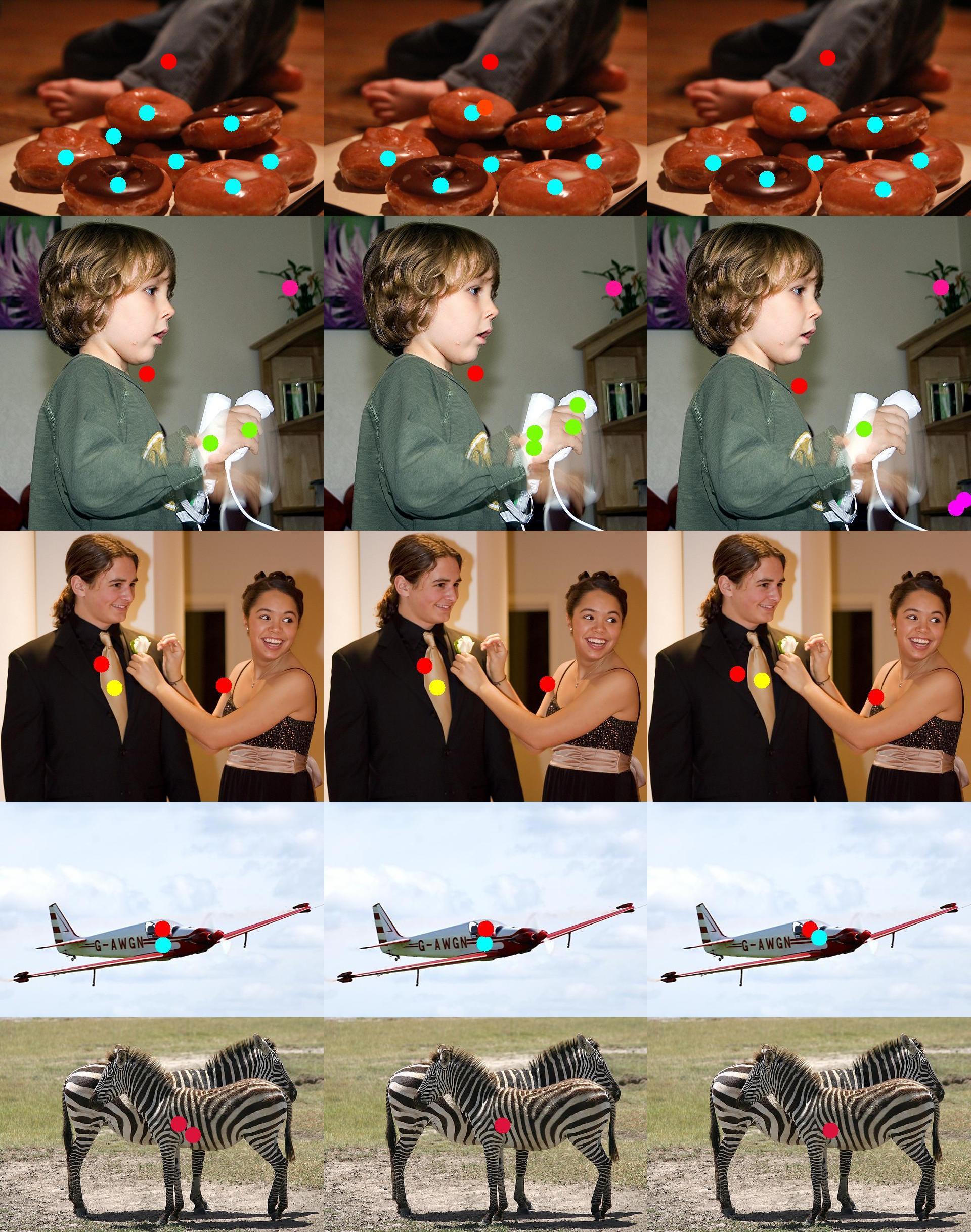}
    \begin{minipage}{0.33\linewidth}
        \centering
        Ground Truth
    \end{minipage}%
    \begin{minipage}{0.33\linewidth}
        \centering
        YOLO11x
    \end{minipage}%
    \begin{minipage}{0.33\linewidth}
        \centering
        OCDet-X (ours)
    \end{minipage}
    \caption{Failure cases of OCDet-X (right) in object center detection compared with Ground Truth (left) and YOLO11x (center). The first two rows display missed detections of less prominent objects. The third and fourth rows show deviations in detected center localization, and the final row presents cases where only one center is detected for overlapping objects.}
    \label{fig:comparison_failure}
\end{figure*}

\begin{figure*}[ht]
    \centering
    \includegraphics[width=0.92\linewidth]{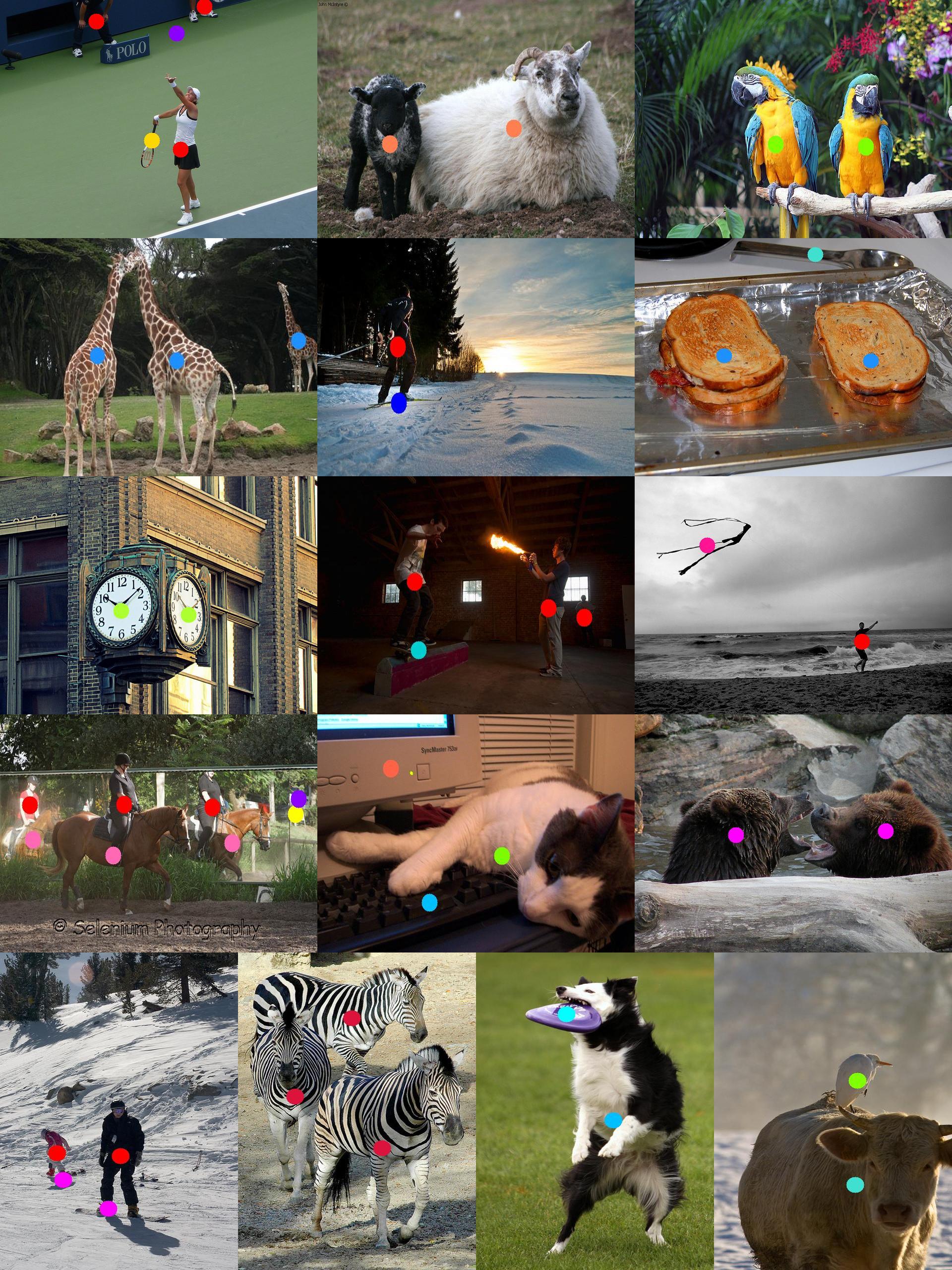}
    \caption{More visualizations of OCDet-X's object center detection results.}
    \label{fig:image_collage}
\end{figure*}

\cref{fig:comparison}, \cref{fig:comparison_failure}, and \cref{fig:image_collage} illustrate qualitative results for object center detection with OCDet. Specifically, \cref{fig:comparison} contrasts the detection results of YOLO11x and OCDet-x against ground-truth annotations. Notably, OCDet accurately identifies object centers even when ground-truth annotations are incorrect (rows 3-6). The strength of OCDet is particularly evident in the final two rows, where it correctly detects significantly more target objects in complex and cluttered scenes than YOLO11. However, OCDet is not without limitations. As shown in \cref{fig:comparison_failure}, OCDet occasionally misses less salient object centers in challenging scenes, especially when YOLO11 also performs poorly (rows 1–2). Although OCDet significantly outperforms YOLO11 in recall, it exhibits comparatively lower precision in localizing centers of easily detectable objects (rows 3–4). For objects with overlapping centers (row 5), both models show limited effectiveness. Finally, \cref{fig:image_collage} provides a comprehensive visualization of OCDet’s performance in detecting object centers across multiple classes.

